\documentclass[a4paper,fleqn]{cas-dc}

\usepackage[authoryear]{natbib}

\usepackage[subrefformat=parens]{subcaption}
\renewcommand{\figurename}{Fig.}

\usepackage{lineno}
\newcommand*\patchAmsMathEnvironmentForLineno[1]{%
  \expandafter\let\csname old#1\expandafter\endcsname\csname #1\endcsname
  \expandafter\let\csname oldend#1\expandafter\endcsname\csname end#1\endcsname
  \renewenvironment{#1}%
     {\linenomath\csname old#1\endcsname}%
     {\csname oldend#1\endcsname\endlinenomath}}%
\newcommand*\patchBothAmsMathEnvironmentsForLineno[1]{%
  \patchAmsMathEnvironmentForLineno{#1}%
  \patchAmsMathEnvironmentForLineno{#1*}}%
\AtBeginDocument{%
\patchBothAmsMathEnvironmentsForLineno{equation}%
\patchBothAmsMathEnvironmentsForLineno{align}%
\patchBothAmsMathEnvironmentsForLineno{flalign}%
\patchBothAmsMathEnvironmentsForLineno{alignat}%
\patchBothAmsMathEnvironmentsForLineno{gather}%
\patchBothAmsMathEnvironmentsForLineno{multline}%
}

\usepackage{ulem}
\newcommand\Erase{\bgroup\markoverwith{\textcolor{red}{\rule[.5ex]{2pt}{0.4pt}}}\ULon}
\newcommand{\Add}[1]{\textcolor{black}{#1}}	%

\newcommand\EraseTwo{\bgroup\markoverwith{\textcolor{magenta}{\rule[.5ex]{2pt}{0.4pt}}}\ULon}

\usepackage{amsmath}   
\usepackage{mathtools} 
\usepackage{cleveref}  

\usepackage{bm}

\usepackage{algorithm}
\usepackage{algpseudocode}

\ExplSyntaxOn
\keys_set:nn { stm / mktitle } { nologo }
\ExplSyntaxOff

\crefname{equation}{Eq.}{Eqs.}%
\crefname{figure}{Fig.}{Figs.}%
\crefname{table}{Tab.}{Tabs.}
\crefformat{section}{Sec.#2#1#3}

\def\tsc#1{\csdef{#1}{\textsc{\lowercase{#1}}\xspace}}
\tsc{WGM}
\tsc{QE}
\tsc{EP}
\tsc{PMS}
\tsc{BEC}
\tsc{DE}

\newcommand{\markupdraft}[2]{
    \ifthenelse{\equal{#1}{display}}{#2}{}
    \ifthenelse{\equal{#1}{color}}{\color{#2}}{}
}

\begin{document}
\let\WriteBookmarks\relax
\def\floatpagepagefraction{1}
\def\textpagefraction{.001}
\shorttitle{Ship trajectory planning method for reproducing human operation at ports}
\shortauthors{Suyama et~al.}

\title [mode = title]{Ship trajectory planning method for reproducing human operation at ports}                      



\author[1]{Rin Suyama}
\cormark[1]
\credit{Conceptualization, Methodology, Software, Validation, Formal analysis, Data Curation, Writing - Original Draft, Visualization}
\address[1]{Department of Naval Architecture and Ocean Engineering, Graduate School of Engineering, Osaka University, 2-1 Yamadaoka, Suita, Osaka 565-0971, Japan}
\ead{suyama_rin@naoe.eng.osaka-u.ac.jp}

\author[1]{Yoshiki Miyauchi}
\credit{Conceptualization, Methodology, Software, Validation, Writing - Review \& Editing}
\ead{yoshiki_miyauchi@naoe.eng.osaka-u.ac.jp}


\author[1]{Atsuo Maki}
\credit{Writing - Review \& Editing, Supervision, Project administration, Funding acquisition}
\cormark[1]
\ead{maki@naoe.eng.osaka-u.ac.jp}

\cortext[cor1]{Corresponding author}

\begin{abstract}
    Among ship maneuvers, berthing/unberthing maneuvers are one of the most challenging and stressful phases for captains.
    Concerning burden reduction on ship operators and preventing accidents, several researches have been conducted on trajectory planning to automate berthing/unberthing.
    However, few studies have aimed at assisting captains in berthing/unberthing.
    The trajectory to be presented to the captain should be a maneuver that reproduces human captain's control characteristics.
    The previously proposed methods cannot explicitly reflect the motion and navigation, which human captains pay particular attention to reduce the mental burden in the trajectory planning.
    Herein, mild constraints to the trajectory planning method are introduced.
    The constraints impose certain states (position, bow heading angle, ship speed, and yaw angular velocity), to be taken approximately at any given time.
    The introduction of this new constraint allows imposing careful trajectory planning (e.g., in-situ turns at zero speed or a pause for safety before going astern), as if performed by a human during berthing/unberthing.
    The algorithm proposed herein was used to optimize the berthing/unberthing trajectories for a large car ferry.
    The results show that this method can generate the quantitatively equivalent trajectory recorded in the actual berthing/unberthing maneuver performed by a human captain.
\end{abstract}




\maketitle

\sloppy

\section{Introduction}

    \label{sec:intro}
    
    Ship maneuver in a narrow and congested port is one of the most challenging operations, and especially, berthing/unberthing maneuver requires high operational skills  \citep{Seta2004,Hasegawa1993_maneuver}.
    When ships are operated in ports, they are navigated at slower speeds than in the open sea.
    Therefore, the rudder force is less effective, and the effects of wind and other external disturbances are more pronounced \citep{Seta2004,Hasegawa1993_maneuver}.
    Thus, berthing/unberthing are considered the most stressful phases of ship operation for inexperienced and highly experienced captains \citep{Seta2004,Hasegawa1993_maneuver}.
    Additionally, the burden will be even heavier if the captain is unfamiliar with the port.
    Therefore, if the ship operator has expected berthing/unberthing trajectories before maneuvering, their burden could be reduced.
    Furthermore, accident risks could also be reduced.
    Such information may include speed profiles around target berthed positions, trajectory plans, and guidelines for controlling actuators, such as rudders and main engines.
    Herein, ``trajectory'' represents the feasible time series of ship's state; position, heading, velocities, and yaw angular velocity. 
    In other words, the trajectory satisfies kinematic and dynamic constraints.
    Trajectory planning/generation entails optimizing a ship's trajectory considering the dynamical performance of the target ship.
    Recently, studies have been conducted extensively for ship operation assistance, and application as a reference trajectory in autonomous ship operation.
    
    The trajectory planned for supporting captains should have the same characteristics as that of human captain because the planned trajectory must be acceptable from captain's perspective.
    Herein, the term ``recommended trajectory'' represents the trajectory planned for supporting the maneuver of the captain.
    The human captain's maneuver has typical characteristics, whereas captains adhere to the rules of the course.
    For instance,
    \begin{itemize}
        \item Stopping in front of the berth at a specific position and heading angle
        \item Choosing a course that maximizes clearance with the surroundings as much as possible
    \end{itemize}
    According to \citet{Koyama1987}, when berthing, first, the captain stops the ships parallel to the berthing line at the front of the berth.
    \citet{Koyama1987} reported that this maneuver reduces the load on the operator.
    The reason for this was analyzed by them as follows: the stopping accuracy required for stopping maneuver in front of a berth is relatively relaxed.
    Additionally, the maneuver from a first stop position to the target berthed position is simpler and less difficult than that required for berthing the ship directly at the target berthed position without any intermission.
    Needless to say, a recommended trajectory can also be used as a reference trajectory for autonomous berthing/unberthing maneuver, which is one of the main objectives of trajectory planning.
    
    So far, research studies on ship berthing/unberthing trajectory planning methods have been conducted mainly as a way of offering reference trajectories for automating berthing/unberthing maneuvers.
    For example, \citet{Martinsen2019} formulated the autonomous berthing problem as an optimal control problem in the framework of model predictive control and the numerically obtained solution.
    Also, \citet{Bergman2020} and \citet{Martinsen2021} proposed a two-step procedure.
    The first candidate trajectory was made, subsequently, it was improved based on optimality.
    Furthermore, \citet{Bitar2020} considered maneuver energy and performed trajectory planning in ports with external disturbances.
    
    In research studies on autonomous ship berthing/unberthing, some methods can be applied to recommended trajectory.
    \citet{Liu2019} proposed an application of the A-star algorithm for optimizing the berthing path.
    They simultaneously considered the path length, collision avoidance, separation distance from surrounding structures, own ship's maneuverability, and currents.
    \citet{Han2021} presented a real-time motion planning method for automatic berthing using Extended Dynamic Window Approach.
    In Japan, since the late 1980s, several research studies on autonomous berthing/unberthing have been conducted \citep{Kose1986,Koyama1987}.
    \citet{Shoji1992,Shoji1992_2,Shoji1993,Shoji1993_2} presented the pioneering works in which the optimal control theory was applied for obtaining the berthing trajectories.
    They formulated the ship autonomous operation problem as a minimum time problem and numerically solved the problem using an indirect method based on the variational method, which is named as Sequential Conjugate Gradient-Restoration Algorithm (SCGRA) \citep{A.K.1980,A.K.1980_2}.
    From this point of view, his works were epoch-making.
    \citet{Maki2020,Maki2020_2} also formulated the optimal control problem of autonomous berthing as an extension of the works of \citet{Shoji1992}.
    The most significant difference from the works of Shoji is the inclusion of the risk of collision to berth.
    This inclusion has never been conducted in almost all previous works except for \citep{Koyama1987}). 
    \citet{Miyauchi2022} further extended the work of \citet{Maki2020,Maki2020_2}.
    They considered the actual port topography by struggling to reformulate the ship domain by considering the wind disturbance.
    Also, they improved the algorithm by introducing waypoints to obtain a favorable trajectory plan in the complex port topography.
    
    Besides those focused on berthing/unberthing maneuver, trajectory/path planning methods exist for generating reference trajectories for automation of ship operation.
    As an application of reinforcement learning, \citet{Amendola2020} investigated the acquisition of control laws for narrow channel maneuver.
    Their method provided solutions that could be used as a trajectory plan.
    
    However, to the best of our knowledge, few studies on recommended trajectories have been presented.
    Studies on berthing/unberthing trajectory planning methods only focused on ship control automation have been progressing.
    Consequently, it is possible to obtain a trajectory plan to complete the maneuver objective safely and in the minimum time.
    However, the method presented so far did not explicitly include constraints to realize a specific state in the trajectory.
    Thus, no research has been conducted to reproduce human berthing/unberthing operations.
    
    Herein, a mild constraint is newly introduced in the planning of the recommended trajectory.
    This constraint imposes the control characteristic of the human operator to be mildly satisfied upon the recommended trajectory.
    The newly introduced constraint condition is referred to as ``checkpoint'' condition.
    ``Waypoint'' is a common term that represents the constraint imposed on the ship's position in the field of ship navigation problems.
    The new checkpoint condition introduced herein is similar to the waypoint at first glance.
    However, it is novel because it constraints the  position, heading angle, ship speed, and yaw angular velocity at the same time.
    Also, our new method can possibly retain the characteristics of a minimum time problem, as treated by \citet{Shoji1992,Maki2020,Miyauchi2022}, because the checkpoint constraint is not strict, but has tolerance on its satisfaction.
    The checkpoint condition could be easily determined by operators with sufficient level of berthing/unberthing maneuver skills.
    
    Herein, we newly propose an algorithm to optimize the trajectory that additionally satisfies the checkpoint condition 
    Here, the algorithm proposed by \citet{Miyauchi2022} is extended.
    In the algorithm proposed herein, a new checkpoint condition is considered in the objective function.
    Then, the state transitions are optimized to satisfy the checkpoint condition at any time.
    Herein, numerical experiments are conducted to verify the proposed algorithm.
    In the numerical experiments, the checkpoint conditions are set based on the recorded maneuvering data of a $222.5 \mathrm{[m]}$-long RoPax vessel.
    We compare the results with the actual operation data and the trajectory generated by the present algorithm and the previous algorithm reported by \citet{Miyauchi2022}.
    Next, the effectiveness of the new checkpoint condition on the berthing/unberthing trajectory planning is evaluated.

\section{Trajectory planning algorithm}

    \subsection{Ship maneuvering system}
    
\begin{figure}[tb]
            \centering
            \includegraphics[width=1.0\hsize]{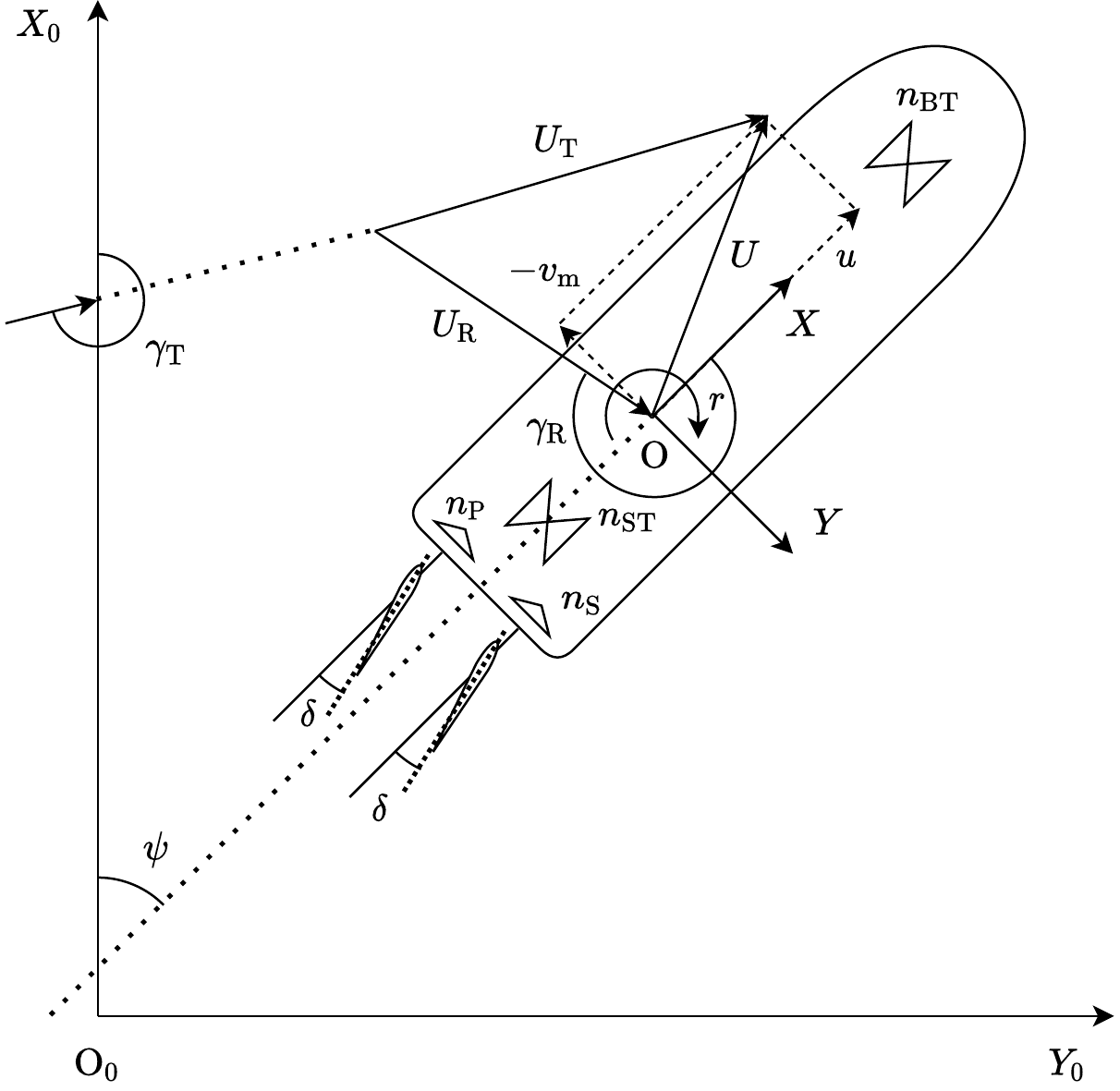}
            \caption{Coordinate systems.}
            \label{fig:coordinate}
        \end{figure}

        The ship was assumed to move on a two-dimensional plane (\Cref{fig:coordinate}) through a duration $0 \leq t \leq t_{\mathrm{f}}$.
        Here $t \in \mathbb{R}$ represents the time.
        In this study, the terminal time $t_{\mathrm{f}} \in \mathbb{R}$ was the variable to be optimized, as explained later.
        $\mathrm{O}_{0}-X_{0}Y_{0}$ in \Cref{fig:coordinate} is the earth-fixed coordinate system, and $\mathrm{O}-XY$ is the ship-fixed coordinate system with the midship as the origin.
        
        The ship's state $\bm{x}(t)$ was defined by \Cref{eq:x}.
        \begin{equation}
            \label{eq:x}
            \bm{x}(t) := ( X_{0}(t), u(t), Y_{0}(t), v_{\mathrm{m}}(t), \psi(t), r(t) ) ^ \top \in \mathbb{R} ^ 6
        \end{equation}
        State $\bm{x}(t)$ comprises the earth-fixed coordinate component $X_{0}(t)$, surge velocity $u(t)$, the earth-fixed coordinate component $Y_{0}(t)$, sway velocity $v_{\mathrm{m}}(t)$, heading angle $\psi(t)$, and yaw angular velocity $r(t)$.
        Each positive direction was set to be the direction indicated in \Cref{fig:coordinate}.
        
\begin{figure}[tb]
            \centering
            \includegraphics[width=1.0\hsize]{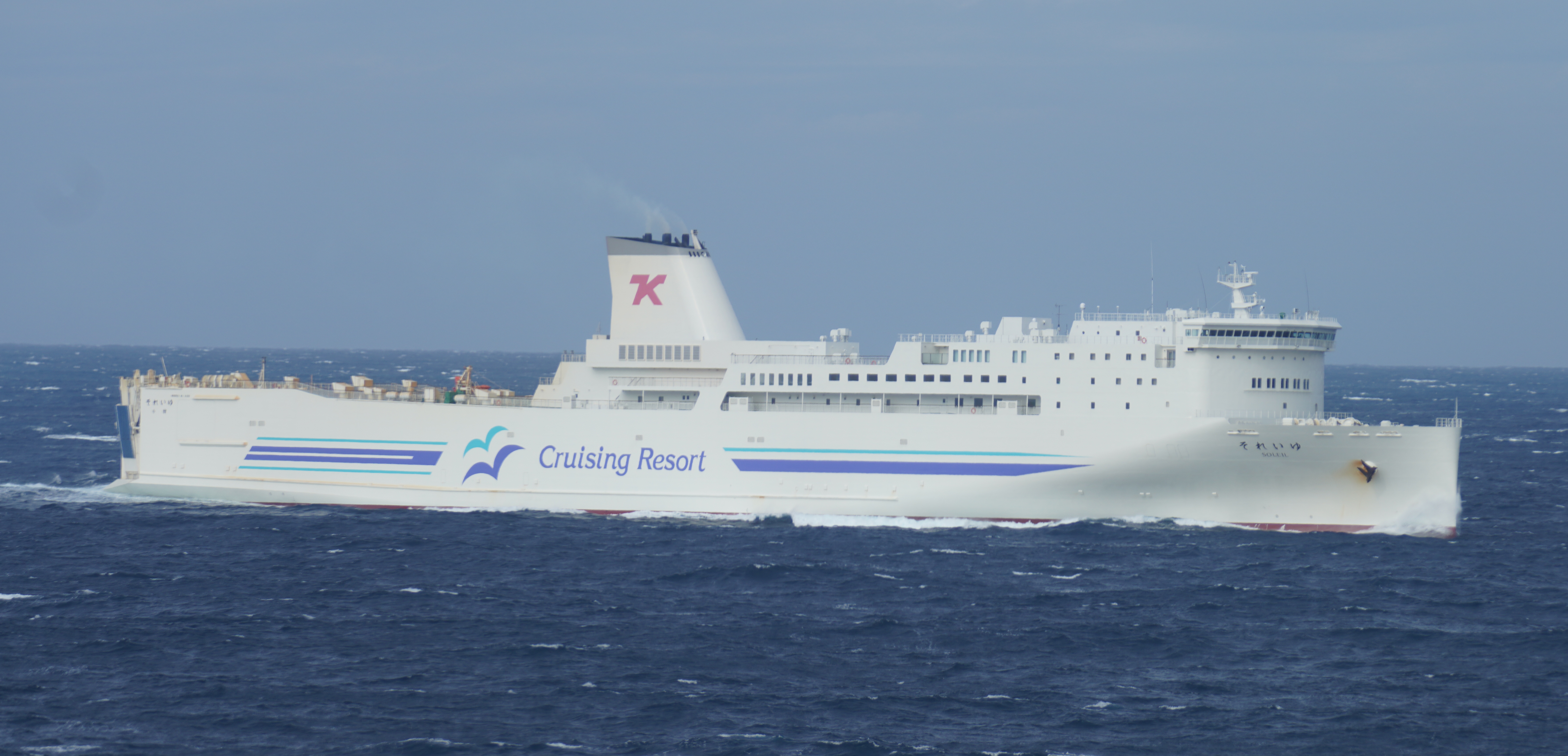}
            \caption{Photo of subject ship; car ferry equipped with two Controllable Pitch Propellers (CPPs), two rudders, bow thrusters, and stern thrusters.}
            \label{fig:ship_photo}
        \end{figure}

        The actuator control input was denoted as $\bm{u}(t)$.
        Herein, a $222.5 \mathrm{[m]}$-long RoPax vessel was used as the subject ship.
        It was equipped with twin propellers having Controllable Pitch Propellers (CPPs) mechanism, twin rudders, bow thrusters, and stern thrusters.
        \Cref{fig:ship_photo} presents the photo of the target ship.
        Herein, the port and starboard rudder angles were always assumed to be equal, so the control input $\bm{u}(t)$ was set to be five-dimensional.
        Although the subject ship is equipped with CPPs,
        the propellers were modeled as fixed pitch propellers (FPPs) in the maneuvering simulation because the parameters of the maneuvering model were generated from the test results of the model ship with FPP.
        However, for computing the results shown in \Cref{sec:results}, the FPP revolution was converted to equivalent CPP pitch angle, which generates the equivalent propeller thrust with the FPP.
        Considering the above, we finally defined $\bm{u}(t)$ by \Cref{eq:u}:
        \begin{equation}
            \label{eq:u}
            \bm{u}(t) := ( \delta(t), n_{\mathrm{P}}(t), n_{\mathrm{S}}(t), n_{\mathrm{BT}}(t), n_{\mathrm{ST}}(t) ) ^ \top \in \mathbb{R} ^ 5 \enspace .
        \end{equation}
        Here, $\delta(t)$ is the rudder angle, $n_{\mathrm{P}}(t)$ and $n_{\mathrm{S}}(t)$ are the revolution numbers of the port and starboard side propellers, and $n_{\mathrm{BT}}(t)$ and $n_{\mathrm{ST}}(t)$ are the revolution numbers of the bow and stern thrusters, respectively.
        In our numerical experiment, multisided thrusters at the fore and aft are assumed to be single thrusters at each.
   
        The maneuvering motion equation for the subject ship was as follows:
        \begin{equation}
            \label{eq:state_eq}
            \dot{\bm{x}}(t) = f_{\mathrm{MMG}} ( \bm{x}(t), \bm{u}(t), \bm{w}(t) )
            \enspace .
        \end{equation}
        The over dot on a variable represents the differentiation with respect to time $t$.
        Here, $\bm{w}(t)$ represents the wind disturbance, which is presented as follows:
        \begin{equation}
            \label{eq:w}
            \bm{w}(t) := ( U_{\mathrm{R}}(t), \gamma_{\mathrm{R}}(t) ) ^ \top \in \mathbb{R} ^ 2
            \enspace ,
        \end{equation}
        where $U_{\mathrm{R}}(t)$ and $\gamma_{\mathrm{R}}(t)$ are the relative wind speed and direction, respectively.
        As shown in \Cref{eq:state_eq}, the right side of the state equation can be calculated for the instantaneous state variable, control input, and wind according to $f_{\mathrm{MMG}} : \mathbb{R} ^ 6 \times \mathbb{R} ^ 5 \times \mathbb{R} ^ 2 \rightarrow \mathbb{R} ^ 6$.
        The hydrodynamic forces and moments considered in $f_{\mathrm{MMG}}$
        were calculated by the \Add{Mathematical Modeling Group (MMG)} model (e.g., \citep{Ogawa1978}).
        \Add{
        Total hydrodynamic forces acting on a ship are divided into segments in the MMG model:
        \begin{equation}
            \label{eq:MMG_forces}
            \left \{
                \begin{aligned}
                    \mathcal{X} &= \mathcal{X}_{\mathrm{H}} + \mathcal{X}_{\mathrm{P}} + \mathcal{X}_{\mathrm{R}} + \mathcal{X}_{\mathrm{S}}  \\
                    \mathcal{Y} &= \mathcal{Y}_{\mathrm{H}} + \mathcal{Y}_{\mathrm{P}} + \mathcal{Y}_{\mathrm{R}} + \mathcal{Y}_{\mathrm{S}}  \\
                    \mathcal{N} &= \mathcal{N}_{\mathrm{H}} + \mathcal{N}_{\mathrm{P}} + \mathcal{N}_{\mathrm{R}} + \mathcal{N}_{\mathrm{S}}
                \end{aligned}
            \right .
        \end{equation}
        Here, $\mathcal{X}, \mathcal{Y}, \mathcal{N}$ are elements of hydrodynamic forces acting on a ship in the $X, Y$ directions, and hydrodynamic yaw moment, respectively.
        The letters $H, P, R$, and $S$ stand for hull, propeller, rudder, and side thruster, respectively.
        A detail of modelling of each term in \Cref{eq:MMG_forces} herein can be found in the literature \citet{Kobayashi1984,Asai1984}.
        }
        The numerical scheme to solve \Cref{eq:state_eq} was the fourth-order Runge--Kutta method.
        Herein, time width $\Delta t_{\mathrm{S}} = 1.0 ~ \mathrm{[s]}$ was chosen.
    
    \subsection{Optimization problem}
        
        \label{sec:optimal_control_problem_formulation}
        
        We formulated the trajectory planning problem in the framework of the optimal control theory, as stated in many previous studies \citep{Shoji1992,Shoji1992_2,Shoji1993,Shoji1993_2,Martinsen2019,Maki2020,Maki2020_2,Bitar2020,Bergman2020,Martinsen2021,Miyauchi2022}.
        Our formulation is shown in this section.
    
        \subsubsection{Formulation}
            
            \label{sec:formulation}
        
            The optimal control theory framework aims to determine the control and state variable sets that minimize or maximize the objective function.
            Generally, the initial and terminal states are given since the problem to be solved is the berthing/unberthing trajectory planning problem.
            Additionally, in the optimal control theory framework, the state equation and the constraints variables are treated as equality and inequality constraints.
            Thus, the problem of automating the berthing/unberthing trajectory planning can be mathematically described as a nonlinear two-point boundary value problem (TPBVP) with equality and inequality constraints.
        
            The formulation detail as an optimal control problem is as follows.
            First, the state equation is rewritten, by defining a new function $\phi : \mathbb{R} ^ 6 \times \mathbb{R} ^ 6 \times \mathbb{R} ^ 5 \times \mathbb{R} ^ 2 \rightarrow \mathbb{R} ^ 6$, as follows:
            \begin{equation}
                \begin{aligned}
                    & \phi( \bm{x}(t), \dot{\bm{x}}(t), \bm{u}(t), \bm{w}(t) )  \\
                    & := f_{\mathrm{MMG}} ( \bm{x}(t), \bm{u}(t), \bm{w}(t) ) - \dot{\bm{x}}(t)  \\
                    & = \bm{0}
                    \enspace .
                \end{aligned}
            \end{equation}
            Herein, the wind disturbance $\bm{w}(t)$ was considered as constant through $0 \leq t \leq t_{\mathrm{f}}$.
            Besides, the boundary conditions for the initial state $\bm{x}(0)$ and terminal state $\bm{x}(t_{\mathrm{f}})$ were set in the form of:
            \begin{equation}
                \left \{
                    \begin{aligned}
                        x_{i}(0) &= x_{\mathrm{init}, i}  \\
                        x_{i}(t_{\mathrm{f}}) &= x_{\mathrm{fin}, i}
                    \end{aligned}
                \right.
                \qquad \text{for} \quad i = 1, \dots, 6
                \enspace ,
            \end{equation}
            where $x_{\mathrm{init}, i}$, $x_{\mathrm{fin}, i}$ are the values at which the $i$th component of $\bm{x}$ at the initial and terminal times should be taken.
            For the control input, the constraints which have the form of:
            \begin{equation}
                g_{j}(u_{j}(t)) \leq 0 \qquad \text{for} \quad j = 1, \dots, 5
                \enspace ,
            \end{equation}
            \Add{where $g_{j} : \mathbb{R} \rightarrow \mathbb{R}$,} were imposed to account for the limit of the actuators.
            \Add{For instance, the constraint on rudder angle $u_{1} = \delta ~ \mathrm{[deg]}$ is formulated as: $g_{1}(u_{1}) := | u_{1} - 35 |$, which corresponds to \Cref{tab:exploring_range}.}
            Similarly, the constraint on the terminal time $t_{\mathrm{f}}$ is formulated as:
            \begin{equation}
                g_{t_{\mathrm{f}}}(t_{\mathrm{f}}) \leq 0
                \enspace \Add{,}
            \end{equation}
            \Add{using a function $g_{t_{\mathrm{f}}} : \mathbb{R} \rightarrow \mathbb{R}$.}
            Additionally, the objective function $I$ to be minimized is usually set in the following form:
            \begin{equation}
                \label{eq:I}
                I := \int_{0}^{t_{\mathrm{f}}} h( \bm{x}(t), \bm{u}(t), t ) \mathrm{d}t
                \enspace .
            \end{equation}
            The objective function reflects the purpose of the control to be achieved, and so far the problems have been formulated as minimum time, shortest path, and minimum energy as reported by \citet{Shoji1992,Shoji1992_2,Shoji1993,Shoji1993_2}.
            
            The optimal control problem described above is generally summarized as shown below.
            \begin{equation}
                \label{eq:ocp}
                \begin{aligned}
                    &\underset{ \bm{u}(t), t_{\mathrm{f}} }{ \text{minimize} } \quad I  \\
                    &\text{s.t.} \quad
                        \left \{
                            \begin{aligned}
                                \Add{\phi}( \bm{x}(t), \dot{\bm{x}}(t), \bm{u}(t), \bm{w}(t) ) &= 0  \\
                                \bm{w}(t) &= \text{Specified.}  \\
                                x_{i}(0) &= x_{\mathrm{init}, i}  \\
                                x_{i}(t_{\mathrm{f}}) &= x_{\mathrm{fin}, i}  \\
                                g_{j}(u_{j}(t)) &\leq 0 \\
                                g_{t_{\mathrm{f}}}(t_{\mathrm{f}}) &\leq 0  \\
                                \text{for} \quad 
                                \left \{
                                    \begin{aligned}
                                        0 &\leq t \leq t_{\mathrm{f}} \\
                                        i &= 1, \dots, 6  \\
                                        j &= 1, \dots, 5
                                    \end{aligned}
                                \right.
                                &\qquad
                            \end{aligned}
                        \right.
                \end{aligned}
            \end{equation}

        \subsubsection{Solution using Covariance Matrix Adaption Evolutionary Computation (CMA-ES)}
            
            Herein, the Covariance Matrix Adaption Evolutionary Computation (CMA-ES) \citep{Hansen2007,Hansen2014}, which is one of the evolutionary computation methods \citep{Hansen2007}, was used as mathematical programming solver.
            Evolutionary computation generally does not require calculating the gradient of the objective function.
            Therefore, the consideration of the differentiability of the objective function is not necessarily required, which is significant advantage.
            This advantage allows the possibility of optimizing ship motions based on complex maneuvering motion models.
            Several methods exist for solving optimal control problems.
            For example, \citet{Shoji1992} adopted the Sequential Conjugate Gradient-Restoration Algorithm (SCGRA) \citep{A.K.1980,A.K.1980_2} based on the variational method as their numerical solution algorithms.
            In contrast, \citet{Maki2020,Maki2020_2} and \citet{Miyauchi2022} discretized the control sequence in the problem and solved it in the mathematical programming framework.
            They successfully solved the derived mathematical programming problem by CMA-ES.
            The objective functions of the problem had no simple form, unlike the quadratic form, since they considered all minimum time, terminal condition, and intervention condition of the ship domain in objective functions.
            Despite these complexities, they successfully solved the formulated problem with the optimization performance of the CMA-ES.
            
            The numerical algorithm for the berthing/unberthing trajectory planning using the CMA-ES is explained in detail below.
            First, time series of $\bm{u}(t)(0 \leq t \leq t_{\mathrm{f}})$ were \Add{discretized} into $m$ segments $\bm{u}_{k} \in \mathbb{R} ^ {5
            } (k = 1, \dots, m)$ at regular time intervals $\Delta t_{\mathrm{C}}$ (for berthing and unberthing simulation, $\Delta t_{\mathrm{C}} = 100 ~ \mathrm{[s]}$ and $\Delta t_{\mathrm{C}} = 60 ~ \mathrm{[s]}$\Add{, respectively}).
            \Add{
            Except for the last interval interrupted at $t = t_{\mathrm{f}}$, control input is constant for every interval of $\Delta t_{\mathrm{C}}$.
            However, as the result of numerical experiments \Cref{fig:compari_in,fig:compari_out} show, side thrusters are modeled to stop when ship speed is high.
            }
            Next, a set of variables were defined to be optimized (instantaneous control inputs and terminal time) as follows:
            \begin{equation}
                \begin{aligned}
                    \bm{X}
                    :=
                    \Bigl(
                        & t_{\mathrm{f}},
                        \delta_{1}, \dots, \delta_{m},
                        n_{\mathrm{P}, 1}, \dots, n_{\mathrm{P}, m},  \\
                        & n_{\mathrm{S}, 1}, \dots, n_{\mathrm{S}, m},
                        n_{\mathrm{BT}, 1}, \dots, n_{\mathrm{BT}, m},  \\
                        & n_{\mathrm{ST}, 1}, \dots, n_{\mathrm{ST}, m}
                    \Bigr) ^ \top
                    \in \mathbb{R} ^ {5 m + 1}
                    \enspace .
                \end{aligned}
            \end{equation}
            Here, upper and lower limits were set on each of $t_{\mathrm{f}}$, $\delta_{k}$, $n_{\mathrm{P}, k}$, $n_{\mathrm{S}, k}$, $n_{\mathrm{BT}, k}$, $n_{\mathrm{ST}, k} (k = 1, \dots, m)$.
            These limits are shown in \Cref{sec:calc_cond_other}.
            \Add{
            There is a relationship between $\Delta t_{\mathrm{C}}$ and the dimension of the optimization target $\bm{X} \in \mathbb{R}^{5m + 1}$.
            In this case, $m$ is calculated by dividing the upper limit of the searching range for $t_{\mathrm{f}}$ by $\Delta t_{\mathrm{C}}$.
            Thus, a smaller $\Delta t_{\mathrm{C}}$ could give us more smooth and sophisticated maneuvers.
            On the other hand, an increase in the number of variables makes the optimization problem to be solved larger.
            }
            In the CMA-ES framework, at a given iteration, $\bm{X} \in \mathbb{R} ^ {5 m + 1}$ is generated stochastically using its mean vector and covariance matrix.
            These are called candidate solutions.
            Then, an evaluation value is calculated for each candidate solution based on the objective function.
            From those values, the mean vector and covariance matrix of the candidate solutions are updated to generate a new age.
            At each candidate solution update, the initialization range becomes narrower.
            Therefore, the mean vector converges to a certain value as the evaluation value calculation, and candidate solution update continues.
            Similar to previous examples \citep{Maki2020_2,Miyauchi2022}, an algorithm that utilizes a restart strategy \citep{Auger2005} was used here.
            In this strategy, the iteration is restarted from new random candidates if the convergence of the candidate solutions is confirmed.
            
            \Add{
            To avoid frequent change in values, control input for $\Delta t_{\mathrm{C}} ~ \mathrm{[s]}$ was kept constant.
            An application of the pseudospectual method \citet{Ross2012} is another method for obtaining smooth control input, and it may be combined with the proposed method.
            }

    \subsection{Objective function and ship domain}
    
        \label{sec:J_before_ship_domain}
        
        The objective function used herein is an extension of that proposed by \citet{Miyauchi2022}.
        The algorithm is basically formulated as a minimum time problem.
        Significant novelty of their study is the introduction of the ship domain for avoiding contact with the static obstacles and keeping necessary clearance from those.
        
        In the formulation by \citet{Miyauchi2022}, the objective function $J_{\mathrm{B}}$ was represented as follows:
        \begin{equation}
            \label{eq:J_before}
            \begin{aligned}
                J_{\mathrm{B}} =
                    & w_{\mathrm{C}} \cdot C
                    +
                    t_{\mathrm{f}}
                        \cdot \sum_{i = 1}^{6}
                            w_{\mathrm{dim},i}
                            \bigg\{
                                 x_{\mathrm{tol},i} ^ {2}
                                \mathbf{1}_{
                                    \{
                                        | x_{\mathrm{fin},i} - x_{i}( t_{\mathrm{f}} ) |
                                        \le
                                        x_{\mathrm{tol}, i}
                                    \}
                                } \\
                            &+
                                w_{\mathrm{pen}}
                                \left \{
                                    x_{\mathrm{fin},i} - x_{i}( t_{\mathrm{f}} )
                                \right \} ^ {2}
                                \mathbf{1}_{
                                    \{
                                        | x_{\mathrm{fin},i} - x_{i}( t_{\mathrm{f}} ) |
                                        >
                                        x_{\mathrm{tol},i}
                                    \}
                                }
                            \bigg\}
            \end{aligned}
        \end{equation}
        
\begin{figure}
            \centering
            \includegraphics[width=1.0\hsize]{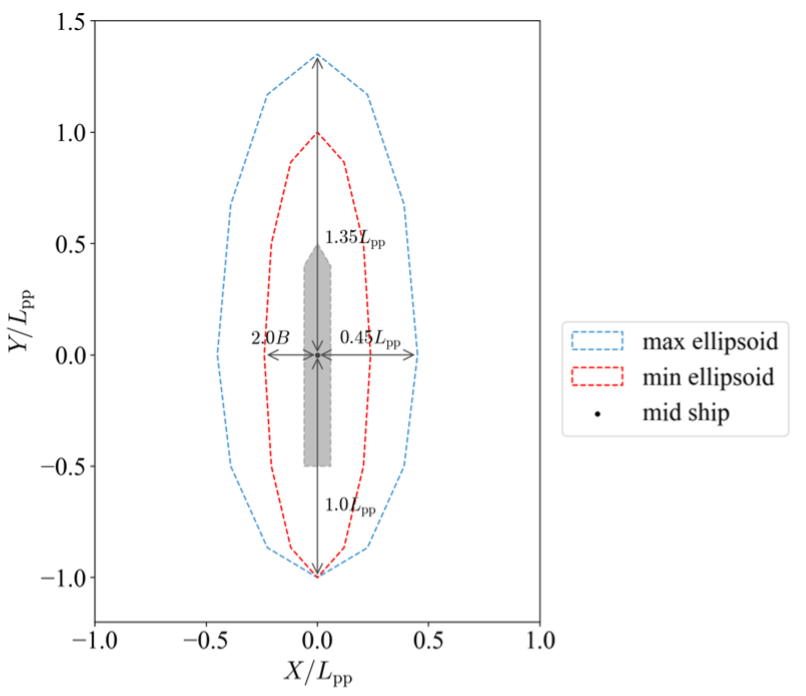}
            \caption{Overview of elliptical shape ship domain.}
            \label{fig:tear_drop_domain}
        \end{figure}

        The first term $w_{\mathrm{C}} \cdot C$ is introduced to avoid collision and maintain trajectory clearance.
        \citet{Miyauchi2022} used the idea of a ship domain \citep{Szlapczynski2017} to avoid collisions between surroundings, such as wharves and breakwaters, and to ensure sufficient separation distance from them.
        The ship domain was a deformed ellipse represented as a dodecagon whose size varies with the ship's speed.
        This formulation was based on modeling of separation distances in ports and harbors \citep{INOUE1994}.
        The ellipse shape was defined by a length of $3 : 2 : 1$ from the center of the ship to the major axis, minor axis, and the ship's side, respectively.
        Here, on the ship-fixed coordinate $X$-axis, the major axis is the direction of the ship's speed $U$, and the minor axis is the opposite direction.
        The ellipse ship domain was set up in the numerical experiments herein, following the method of \citet{Miyauchi2022} (\Cref{fig:tear_drop_domain}).
        For the detailed calculation of the ellipse ship domain, refer to \citet{Miyauchi2022}.
        
        $C \in \mathbb{R}$ in the first term of \Cref{eq:J_before} was the time integral of the sum of the intrusion lengths into the area of surrounding structures at each point of the ship domain.
        $C$ was calculated as follows according to \Cref{eq:C,eq:c_ij}.
        \begin{equation}
            \label{eq:C}
            C :=
            \int_{0}^{t_{\mathrm{f}}}
                \sum_{i_{\mathrm{O}} = 1}^{n_{\mathrm{O}}}
                \sum_{j_{\mathrm{v}} = 1}^{n_{\mathrm{v}}}
                    c_{ i_{\mathrm{O}} j_{\mathrm{v}} }(t)
                \mathrm{d}t
        \end{equation}
        \begin{equation}
            \label{eq:c_ij}
            c_{ i_{\mathrm{O}} j_{\mathrm{v}} }(t)
            :=
            \left \{
                \begin{aligned}
                    L_{ \mathrm{Penet}, i_{\mathrm{O}} j_{\mathrm{v}} }
                        \quad \text{for} \quad
                            P_{j_{\mathrm{v}}}(t) \in S_{\mathrm{O}, i_{\mathrm{O}}}  \\
                    0
                        \quad \text{for} \quad
                            P_{j_{\mathrm{v}}}(t) \notin S_{\mathrm{O}, i_{\mathrm{O}}}
                \end{aligned}
            \right.
        \end{equation}
        In \Cref{eq:C,eq:c_ij}, $L_{ \mathrm{Penet}, i_{\mathrm{O}} j_{\mathrm{v}} }$ is the penetration length of vertex $P_{j_{\mathrm{v}}}(t)$ into the structure region $S_{\mathrm{O}, i_{\mathrm{O}}}$.
        The simulation dealt with the ship motion in a port with a static structure $S_{\mathrm{O}, i_{\mathrm{O}} = 1, \dots, n_{\mathrm{O}}}$ represented by polygonal regions.
        First, a time series of the ship motion was generated according to the control input and the terminal time $\bm{X}$.
        From this time series, a point set $P_{j_{\mathrm{v}} = 1, \dots, n_{\mathrm{v}}}(t)$ representing the ship domain at each time was calculated.
        Here, $n_{\mathrm{v}} =12$ is the total number of vertices of the polygon representing the ship domain.
        Then, according to \Cref{eq:C,eq:c_ij}, the amount of intrusion of each point $P_{j_{\mathrm{v}}}(t)$ into each surroundings structure domain $S_{\mathrm{O}, i_{\mathrm{O}}}$ was calculated.
        \citet{Miyauchi2022} defined $C$ as above and set a sufficiently large constant $w_{\mathrm{C}} > 0$.
        From the above, it can be understood that the first term of \Cref{eq:J_before} introduces a penalty to $J_{\mathrm{B}}$ if the ship domain intrudes into the surrounding structure.
        However, if no interference exists between the ship domain and the surrounding structures in the maneuvering time series produced by $\bm{X}$, $C = 0$, and the first term is zero.
        Herein, $w_{\mathrm{C}} = 1.0 \times 10 ^ {6}$ was used as a sufficiently large value (\Cref{tab:calc_cond_terminal}).
        
        The second term in \Cref{eq:J_before} corresponds to the terminal condition.
        In the approach where the optimal control problem is transformed into a mathematical programming problem, it is difficult to search for a solution that strictly satisfies the terminal condition.
        Therefore, \citet{Miyauchi2022} adopted the method proposed by \citet{Maki2020} and incorporated the terminal condition into the objective function by using the penalty method in the following way.
        First, tolerances were introduced for the terminal conditions.
        The tolerance $\bm{x}_{\mathrm{tol}} \in \mathbb{R} ^ {6}$ for each target terminal state $\bm{x}_{\mathrm{fin}}$ was used as the threshold to satisfy the terminal condition.
        If each deviation component between the terminal state $\bm{x}(t_{\mathrm{f}})$ and $\bm{x}_{\mathrm{fin}}$ falls within the tolerance, the terminal condition was considered satisfied.
        In this situation, the second term of \Cref{eq:J_before} gave a small value.
        However, if the deviation exceeded the tolerance value, the second term in \Cref{eq:J_before} was increased by the large constant $w_{\mathrm{pen}} > 0$.
        Here, the indicator function in the second term is defined with given condition $A$ as follows:
        \begin{equation}
            \mathbf{1}_{A}
            =
            \left \{
                \begin{aligned}
                    & 1
                        \quad & \text{for} \quad
                            & \text{$A$ is satisfied.}  \\
                    & 0
                        \quad & \text{for} \quad
                            & \text{$A$ is not satisfied.} 
                \end{aligned}
            \right.
        \end{equation}
        The vector $\bm{x}_{\mathrm{dim}} \in \mathbb{R} ^ {6}$ was used to nondimensionalize each state in calculating the penalty on the terminal condition satisfaction.
        The settings of $\bm{x}_{\mathrm{tol}}$ and $\bm{x}_{\mathrm{dim}}$ used herein are shown in \Cref{tab:calc_cond_terminal}.
        Additionally, the second term of \Cref{eq:J_before} is the product of the value of the terminal condition and $t_{\mathrm{f}}$, which propels the minimization of operation time of solution.

    \subsection{Point of ship domain transition}
    
        \label{sec:region_switch}
        
\begin{figure}[tb]
            \centering
            \includegraphics[width=1.0\hsize]{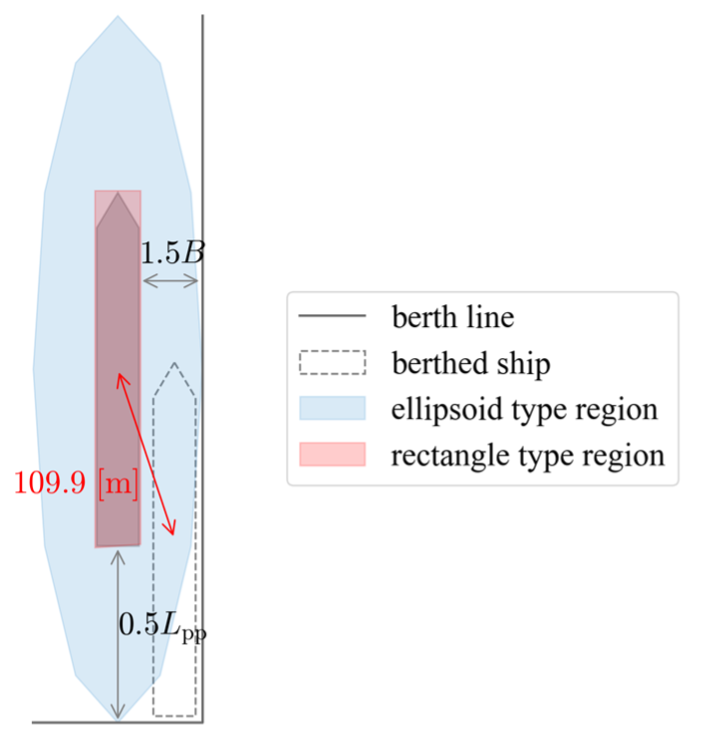}
            \caption{Geometric relation of the ship, ellipsoid type ship domain, rectangular type ship domain, and berth.}
            \label{fig:domain_switch}
        \end{figure}

        In the numerical experiment herein, the algorithm concerning the shape of ship domain was first modified.
        This modification was due to the following reason.
        In the numerical experiment, a target terminal condition was set for the berthing trajectory and the initial condition for the unberthing trajectory close to the berth (\Cref{fig:domain_switch}).
        Obviously, the elliptical-shaped ship domain interferes with the berth in the berthing condition.
        A schematic of these relations is shown in \Cref{fig:domain_switch}.
        Therefore, a new rectangular ship domain was introduced.
        \Cref{fig:domain_switch} shows the newly introduced red-colored rectangular ship domain.
        During the numerical experiments, for the berthing trajectory optimization, the ship domain was an elliptical region initially, but switched to a rectangular ship domain in the vicinity of the target berthed position.
        However, in the case of unberthing trajectory optimization, after the ship attains a sufficient separation distance from the berth, the rectangular ship domain is switched to the elliptical domain.
        The detail of the switching conditions is described later in this section.
        
        The shape of the rectangular region was formulated to have a margin length of $1.0 ~ \mathrm{[m]}$ on each outer side of the circumscribed rectangle of the ship.
        Here, the circumscribed rectangle is of length $L_{\mathrm{pp}}$ times ship width $B$.
        \Add{
        \Cref{fig:overview_rec_and_tol,fig:exact_rec_and_tol} depict the target terminal state, pose deviation due to tolerance, and the rectangular ship domain around the berth.
        \begin{figure}[tb]
            \centering
            \includegraphics[width=1.0\hsize]{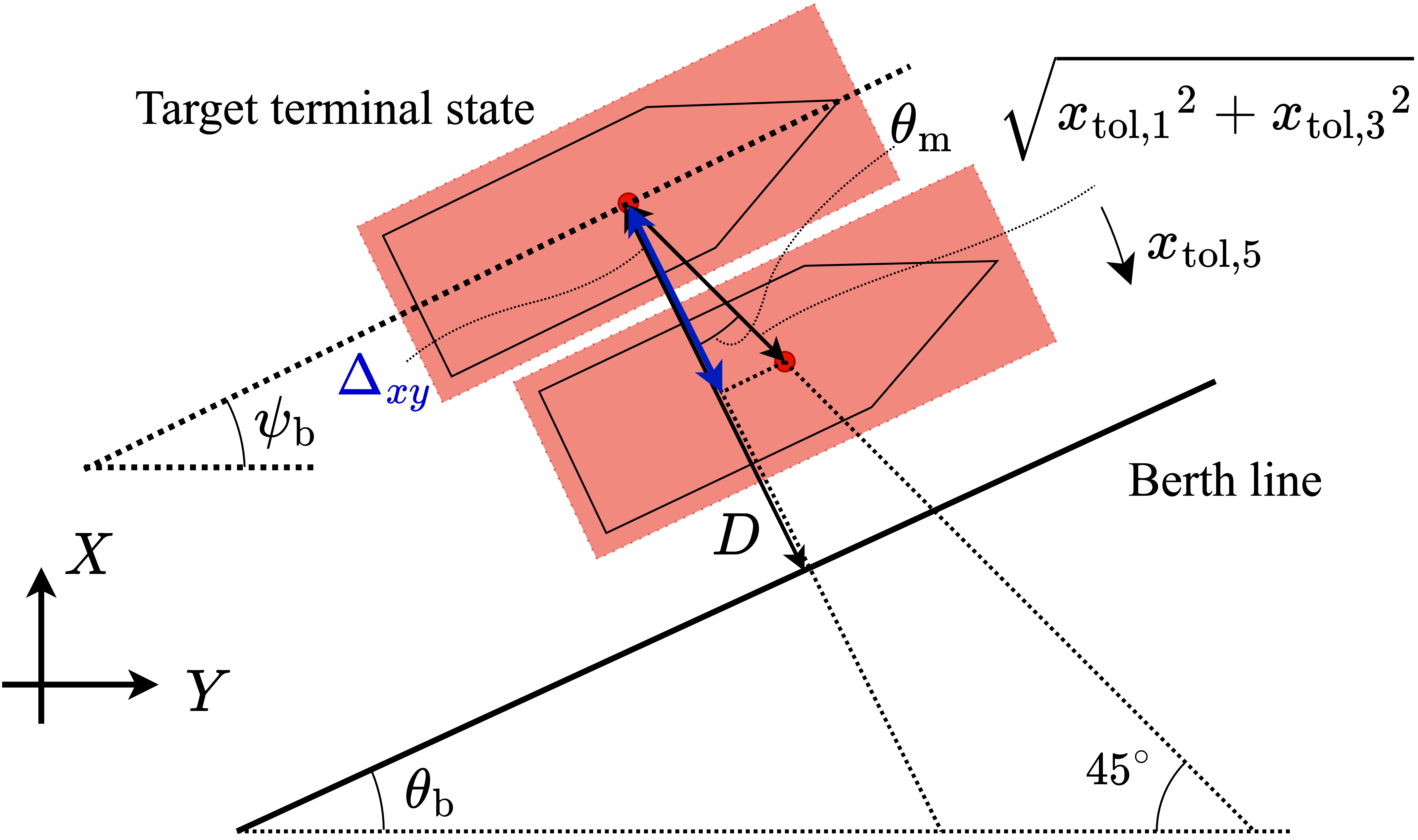}
            \caption{
                \Add{
                Overview of terminal state, rectangular ship domain, and berth line. In this figure, the aspect and positions of the ships are incorrect, and the ships are aligned with the berth line.
                }
            }
            \label{fig:overview_rec_and_tol}
        \end{figure}
        \begin{figure}[tb]
            \centering
            \includegraphics[width=1.0\hsize]{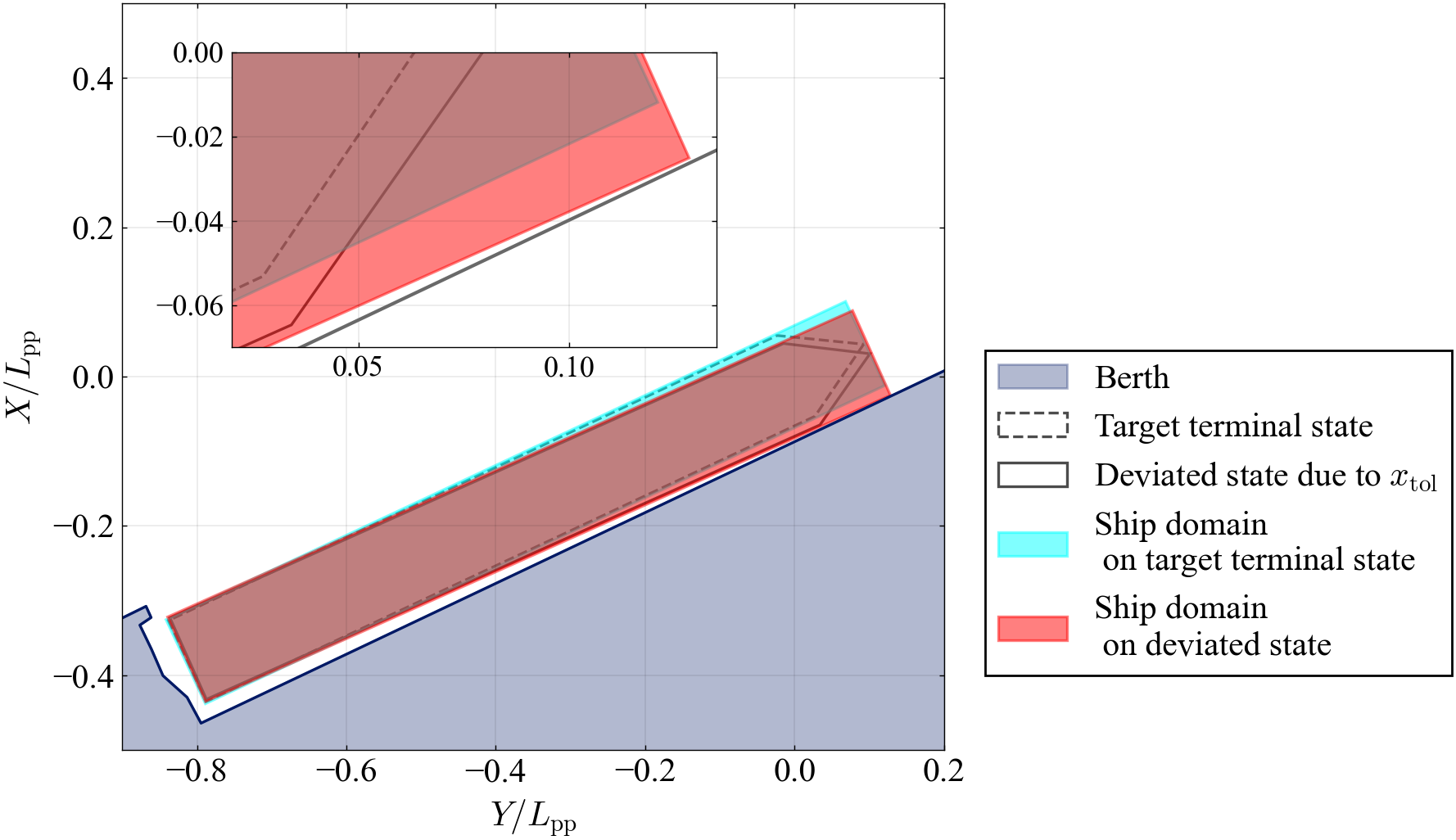}
            \caption{\Add{Exact depiction of terminal state, rectangular ship domain, and berth.}}
            \label{fig:exact_rec_and_tol}
        \end{figure}
        }
        \Add{
        The rectangular ship domain's margin was chosen so that the intervention of the ship domain into the berth is avoided if the target terminal condition we specified is met.
        The target terminal state $\bm{x}_{\mathrm{fin}}$ was set so that the clearance between the starboard side and berth line was $D = 4.0 ~ \mathrm{[m]}$ at midship, and the heading angle was $\psi_{\mathrm{b}} = 25.10 ~ \mathrm{[deg]}$ from the positive side of $Y$ axis.
        In our setting of the tolerance $\bm{x}_{\mathrm{tol}}$, rectangular ship domain with margin lengths of $1.0 ~ \mathrm{[m]}$ were shown not to intrude into the berth region at the terminal state.
        Refer to \Cref{sec:appendix} for more information on this issue.
        }
        
        The region switching was subject to the conditions regarding the distance to the target berthed position and the heading deviation from the berthed attitude.
        The condition regarding the distance to the berthed position was defined with a threshold value of $110 ~ \mathrm{[m]}$.
        This distance was determined by considering the situation near the berthed location (\Cref{fig:domain_switch}).
        A ship surrounded by a minimum elliptical ship domain (light blue area) with the same heading angle as a berthed attitude can approach the berthed position until $109.9 ~ \mathrm{[m]}$ (\Cref{fig:domain_switch}).
        Therefore, the threshold value was set to be $110 ~ \mathrm{[m]}$ in the condition regarding the distance to the berthed position.
        The heading deviation threshold in the ship domain switching condition was $20 ~ \mathrm{[deg]}$.
        According to the consideration of actual maneuver when berthing \citep{Inoue2013}, in an ideal environment, as a rule, a ship approaching the berth moves at an angle of $10 \sim 20 ~ \mathrm{[deg]}$ with the berth line.
        To follow this principle, in berthing trajectory optimization, the ship domain is set to switch from an elliptical shape to a rectangular shape if the heading deviation from the berthed attitude is within $20 ~ \mathrm{[deg]}$.
        In the berthing trajectory planning, the region was set to switch if the conditions related to distance and heading angle were satisfied.
        However, in the unberthing trajectory planning, only the condition regarding distance was used for switching,  because there was no description regarding the heading angle in unberthing maneuver in \citep{Inoue2013}.

    \subsection{Checkpoint condition}
    
        \label{sec:check_point_cond}
        
        %
        This subsection explains ``checkpoint condition.''
        Checkpoint condition imposes inequality constraints on trajectory planning.
        The inequality constraint is on speed $U_{\mathrm{CP}}$, yaw angular velocity $r_{\mathrm{CP}}$, position $(x_{\mathrm{CP}}, y_{\mathrm{CP}})$ and heading angle $\psi_{\mathrm{CP}}$, with gentle tolerances.
        First, a target checkpoint state was specified for each of the four modes of position, heading angle, speed, and yaw angular velocity.
        Second, the inequality constraint was imposed on the deviation between the actual state and the target checkpoint state for each four modes.
        To obtain the recommended trajectory, which satisfies the checkpoint conditions, inequality constraint conditions were incorporated into the objective function in the framework of the penalty method.
        In the penalty method, the imposition of the inequality constraints is replaced by adding the penalty term into the objective function.
        Therefore, the penalty is formulated as the term takes a small value if the constraints are satisfied.
        However, it takes a large value if the constraints are broken.
        We formulated the penalty term on $j$th ($1 \leq j \leq n_{\mathrm{CP}}$, where $n_{\mathrm{CP}}$ is the total number of checkpoint conditions) checkpoint condition $\mathcal{C}_{j}$, which is denoted by $P_{\mathrm{CP}, j}$ (\Cref{eq:P_CP_i}).
        \begin{equation}
            \label{eq:P_CP_i}
            P_{\mathrm{CP}, j} := \min_{i}
                \left (
                    P^{\mathrm{CP, pos}}_{ i j }
                    + P^{\mathrm{CP}, \psi}_{ i j }
                    + P^{\mathrm{CP}, U}_{ i j }
                    + P^{\mathrm{CP}, r}_{ i j }
                \right )
        \end{equation}
        Here, $i$ indicates the timestep progression at every $\Delta t_{\mathrm{S}} ~ \mathrm{[s]}$.
        The detail of \Cref{eq:P_CP_i} is explained later in this section.
        Our new objective function $J(\bm{X}) : \mathbb{R} ^ {5 m + 1} \rightarrow \mathbb{R}$ was defined by adding the sum of the penalties for each checkpoint condition to \Cref{eq:J_before} as follows:
        \begin{equation}
            \label{eq:J}
            J(\bm{X}) = J_{\mathrm{B}} + \sum_{j = 1}^{n_{\mathrm{CP}}} P_{\mathrm{CP}, j}
            \enspace .
        \end{equation}
        
        The detail of \Cref{eq:P_CP_i} is presented as follows.
        In the parenthesis, the sum of each penalty portion for the four modes: position, heading, speed, and yaw angular velocity at each timestep $i$ is calculated.
        Then, $P_{\mathrm{CP}, j}$ is defined by taking the minimum summation value over all timestep $i$.
        Here, each part of \Cref{eq:P_CP_i} was defined by \Cref{alg:P_CP_pos,alg:P_CP_psi,alg:P_CP_U,alg:P_CP_r}.
        \begin{algorithm}[t]
            \caption{Determination of $P^{\mathrm{CP, pos}}_{ij}$}
            \label{alg:P_CP_pos}
            \begin{algorithmic}
                \If{
                    $
                        \sqrt{ ( x_{i} - x_{\mathrm{CP}, j} ) ^ {2} + ( y_{i} - y_{\mathrm{CP}, j} ) ^ {2} }
                        \leq
                        \Add{d_{\mathrm{CP, tol}, j}}
                    $
                }
                    \State{
                        $
                            P^{\mathrm{CP, pos}}_{ij}
                            =
                            \frac
                            { ( x_{i} - x_{\mathrm{CP}, j} ) ^ {2} + ( y_{i} - y_{\mathrm{CP}, j} ) ^ {2} }
                            { d_{\mathrm{CP, tol}, j} ^ {2} }
                        $
                    }
                \Else
                    \State{
                        $
                            P^{\mathrm{CP, pos}}_{ij}
                            =
                            w_{\mathrm{CP, pen}}
                            \frac
                            { ( x_{i} - x_{\mathrm{CP}, j} ) ^ {2} + ( y_{i} - y_{\mathrm{CP}, j} ) ^ {2} }
                            { d_{\mathrm{CP, tol}, j} ^ {2} }
                        $
                    }
                \EndIf
            \end{algorithmic}
        \end{algorithm}
        \begin{algorithm}[t]
            \caption{Determination of $P^{\mathrm{CP}, \psi}_{ij}$}
            \label{alg:P_CP_psi}
            \begin{algorithmic}[0]
                \If{
                    $
                        \Delta \psi_{i}
                        \leq
                        \psi_{\mathrm{CP, tol}, j}
                    $
                }
                    \State{
                        $
                            P^{\mathrm{CP}, \psi}_{ij}
                            =
                            \left (
                                \frac{ \Delta \psi_{i} }{ \psi_{\mathrm{CP, tol}, j}}
                            \right ) ^ {2}
                        $
                    }
                \Else
                    \State{
                        $
                            P^{\mathrm{CP}, \psi}_{ij}
                            =
                            w_{\mathrm{CP, pen}}
                            \left (
                                \frac{ \Delta \psi_{i} }{ \psi_{\mathrm{CP, tol}, j}}
                            \right ) ^ {2}
                        $
                    }
                \EndIf
            \end{algorithmic}
        \end{algorithm}
        \begin{algorithm}[t]
            \caption{Determination of $P^{\mathrm{CP}, U}_{ij}$}
            \label{alg:P_CP_U}
            \begin{algorithmic}[0]
                \If{
                    $
                        | U_{i} - U_{\mathrm{CP}, j} |
                        \leq
                        U_{\mathrm{CP, tol}, j}
                    $
                }
                    \State{
                        $
                            P^{\mathrm{CP}, U}_{ij}
                            =
                            \left (
                                \frac{ U_{i} - U_{\mathrm{CP}, j} }{ U_{\mathrm{CP, tol}, j}}
                            \right ) ^ {2}
                        $
                    }
                \Else
                    \State{
                        $
                            P^{\mathrm{CP}, U}_{ij}
                            =
                            w_{\mathrm{CP, pen}}
                            \left (
                                \frac{ U_{i} - U_{\mathrm{CP}, j} }{ U_{\mathrm{CP, tol}, j}}
                            \right ) ^ {2}
                        $
                    }
                \EndIf
            \end{algorithmic}
        \end{algorithm}
        \begin{algorithm}[t]
            \caption{Determination of $P^{\mathrm{CP}, r}_{ij}$}
            \label{alg:P_CP_r}
            \begin{algorithmic}[0]
                \If{
                    $
                        | r_{i} - r_{\mathrm{CP}, j} |
                        \leq
                        r_{\mathrm{CP, tol}, j}
                    $
                }
                    \State{
                        $
                            P^{\mathrm{CP}, r}_{ij}
                            =
                            \left (
                                \frac{ r_{i} - r_{\mathrm{CP}, j} }{ r_{\mathrm{CP, tol}, j}}
                            \right ) ^ {2}
                        $
                    }
                \Else
                    \State{
                        $
                            P^{\mathrm{CP}, r}_{ij}
                            =
                            w_{\mathrm{CP, pen}}
                            \left (
                                \frac{ r_{i} - r_{\mathrm{CP}, j} }{ r_{\mathrm{CP, tol}, j}}
                            \right ) ^ {2}
                        $
                    }
                \EndIf
            \end{algorithmic}
        \end{algorithm}
        Here, $( x_{\mathrm{CP}, j}, y_{\mathrm{CP}, j} )$, $\psi_{\mathrm{CP}, j}$, $U_{\mathrm{CP}, j}$, $r_{\mathrm{CP}, j}$ indicate position, heading, speed, and yaw angular velocity indicated by the checkpoint condition $\mathcal{C}_{j}$, respectively.
        Besides, $( x_{i}, y_{i} )$, $\psi_{i}$, $U_{i}$, $r_{i}$ are the actual values of position, heading, speed, and yaw angular velocity at timestep $i$.
        In \Cref{alg:P_CP_psi}, $ 0 \leq \Delta \psi_{i} \leq \pi $ represents the subtended angle between the actual heading $\psi_{i}$ and the checkpoint heading $\psi_{\mathrm{CP}, j}$ defined by:
        \begin{equation}
            \label{eq:delta_psi}
            \Delta \psi_{i} := | \Add{\text{atan2}}\{ \sin( \psi_{i} - \psi_{\mathrm{CP}, j} ), \cos( \psi_{i} - \psi_{\mathrm{CP}, j} ) \} |
            \enspace .
        \end{equation}
        As shown in \Cref{alg:P_CP_pos,alg:P_CP_psi,alg:P_CP_U,alg:P_CP_r}, the penalty on each of the four modes was set to be small if the inequality constraint conditions were satisfied.
        However, the penalties take a large value if the conditions are unsatisfied within the influence of $w_{\mathrm{CP, pen}} = 1.0 \times 10 ^ {4} ~ \mathrm{[s]}$.
        
        In \Cref{alg:P_CP_pos,alg:P_CP_psi,alg:P_CP_U,alg:P_CP_r}, these conditions are shown.
        Concerning the position $(x, y)$, the condition was satisfied if the distance between the position $ ( x_{\mathrm{CP}, j}, y_{\mathrm{CP}, j} ) $ designated by the checkpoint condition $ \mathcal{C}_{j} $ and the actual value $ ( x_{i}, y_{i} ) $ was within the tolerance $d_{\mathrm{CP, tol}, j}$.
        Concerning the heading angle $\psi$, the condition was satisfied if the subtended angle between
        $ \psi_{\mathrm{CP}, j} $ designated by the checkpoint condition $ \mathcal{C}_{j} $ and the actual value $ \psi_{i} $ was within the tolerance $\psi_{\mathrm{CP, tol}, j}$.
        Concerning the speed $U$, the condition was satisfied if the difference in speed between
        $U_{\mathrm{CP}, j}$ designated by the checkpoint condition $ \mathcal{C}_{j} $ and the actual value $ U_{i} $ was within the tolerance $U_{\mathrm{CP, tol}, j}$.
        Concerning the yaw angular velocity $r$, the condition was satisfied if the difference in the yaw angular velocity between
        $r_{\mathrm{CP}, j}$ designated by the checkpoint condition $ \mathcal{C}_{j} $ and the actual value $ r_{i} $ was within the tolerance $r_{\mathrm{CP, tol}, j}$.

\section{Numerical experiments}

    \label{sec:num_calc}
    
    Numerical experiments were performed to verify the proposed algorithm.
    In this calculation, checkpoint states were determined by referring to the actual maneuvering data.
    Hereinafter, the terminology ``captain maneuvering data'' is used to represent the actually measured data.
    Then, the performance of the proposed algorithm is validated against the previous algorithm \citep{Miyauchi2022}.

    \subsection{Calculation condition}
    
        \subsubsection{Checkpoint conditions}
            
\begin{figure*}[t]
                \centering
                \includegraphics[width=1.0\hsize]{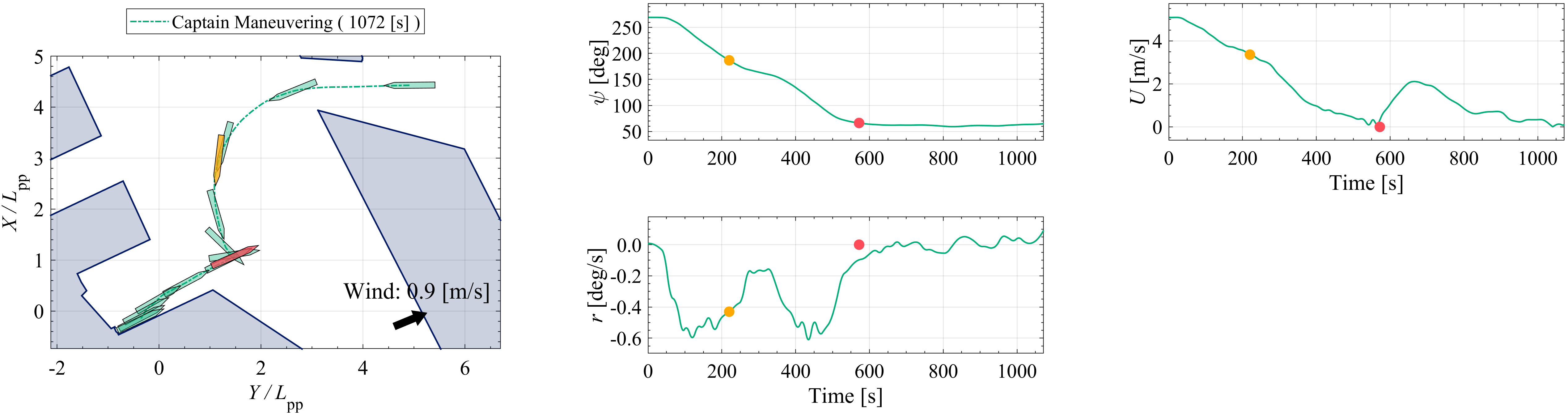}
                \caption{Overview of berthing maneuver in captain maneuvering data. Colored plots are checkpoint states set in our numerical experiment; orange and red represent $\mathcal{C}_{ \mathrm{in}, 1 }$ and $\mathcal{C}_{ \mathrm{in}, 2 }$, respectively.}
                \label{fig:cap_in}
            \end{figure*}

\begin{figure*}[t]
                \centering
                \includegraphics[width=1.0\hsize]{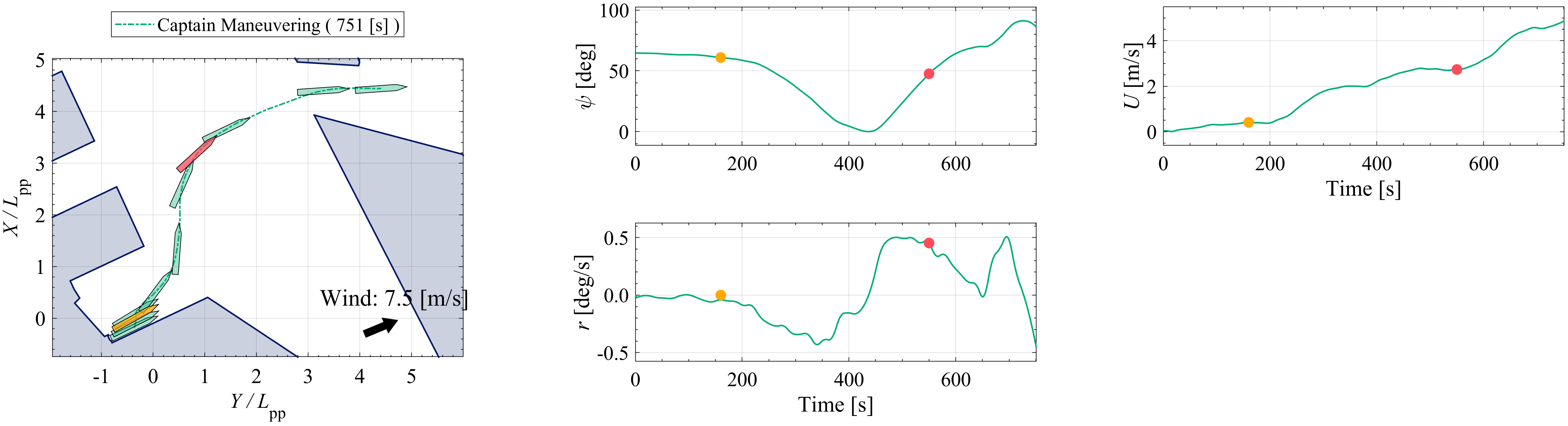}
                \caption{Overview of unberthing maneuver in captain maneuvering data. Colored plots are checkpoint states set in our numerical experiment; orange and red represent $\mathcal{C}_{ \mathrm{out}, 1 }$ and $\mathcal{C}_{ \mathrm{out}, 2 }$, respectively.}
                \label{fig:cap_out}
            \end{figure*}

            The captain maneuvering data includes information on the three degrees of freedom maneuvering motion of the subject ship and measured wind on board.
            For the berthing and unberthing cases, the dataset was from the port entry to the berth and from the berth to the port exit, respectively.
            The data were measured on March 16, 2021, at the Otaru Port, Hokkaido, Japan.
            The measured contents are summarized as follows:
            \begin{itemize}
                \item Time
                \item Latitude and longitude of the bridge position
                \item Heading
                \item Ship speed at the bridge
                \item Direction of ship speed vector at the bridge
                \item Wind speed and direction
            \end{itemize}
            The overview of the captain maneuvering data is summarized in \Cref{tab:cap_data}.
            \begin{table}[h]
                \centering
                \caption{Captain Maneuvering Data}
                \begin{tabular}{|c||c|c|}
                    \hline
                    Item & Berthing & Unberthing  \\
                    \hline
                    \hline
                    Start time & 20:21 & 23:21  \\
                    \hline
                    Duration & $1072 ~ \mathrm{[s]}$ & $751 ~ \mathrm{[s]}$  \\
                    \hline
                    Sampling period & \multicolumn{2}{c|}{About $15 ~ \mathrm{[s]}$}  \\
                    \hline
                    \begin{tabular}{c}
                        Sampling period of  \\
                        wind information  \\
                    \end{tabular}
                    &
                    \multicolumn{2}{c|}{$10 ~ \mathrm{\Add{[minute]}}$}  \\
                    \hline
                    Representative wind speed & $0.9 ~ \mathrm{[m/s]}$ & $7.5 ~ \mathrm{[m/s]}$  \\
                    \hline
                    Representative wind direction & WSW & WSW  \\
                    \hline
                \end{tabular}
                \label{tab:cap_data}
            \end{table}
            The time series data were recorded with a sampling period of approximately $15 ~ \mathrm{[s]}$.
            These time series were interpolated by \Add{spline interpolation} into the time series with a time width of $1.0 ~ \mathrm{[s]}$.
            The captain maneuvering data shown in the figures herein are the interpolated data.
            Meanwhile, the interpolation of the time series of wind information recorded at $10 ~ \mathrm{\Add{[minute]}}$ was not performed.
            The time series of the interpolated captain maneuvering data is depicted in \Cref{fig:cap_in,fig:cap_out}.
            In \Cref{fig:cap_in,fig:cap_out}, the whole trajectory implemented by the captain and time series of $\psi$, $U$ and $r$ are shown.
            In the trajectory diagram, a green-colored dotted line represents the transition of the ship's center.
            The ship's shape is plotted for every $100 ~ \mathrm{[s]}$ from the initial and terminal states.
            The top and bottom of the trajectory diagram correspond to north and south, and the left and right correspond to west and east, respectively.
            In the figures that follow herein, the plots of the shapes of the ship are drawn every $100 ~ \mathrm{[s]}$.
            The navy-colored areas in trajectory diagram are surroundings, such as berth and breakwater, and prohibited areas to enter, such as shallow waters, in the Otaru Port.
            
            We found that there exists a boundary state in the  berthing/unberthing maneuvers in the captain maneuvering data.
            We split the maneuvering motion of the subject ship into two phases by this boundary state.
            In the berthing maneuver (\Cref{fig:cap_in}), we obtained the following features.
            First, after passing the port entrance, the captain turns the ship to the left and navigates while keeping sufficient distance from obstacles.
            On continuous mild turning, the captain stops the ship at approximately $2 L_{\mathrm{pp}}$ from the target berthed position with decreasing speed, and the heading aligns with the target berthed attitude.
            Afterward, the ship is maneuvered to approach the target berthed position at a low speed with a slight heading angle change.
            The above berthing sequence shows that splitting this maneuver into two phases is easy.
            The first phase is transitioning from passing through the port entrance to stopping and turning in front of the berth.
            The second phase is transitioning to approach the target berthed position in a translational motion.
            Similarly, the unberthing maneuver can be divided into two phases, as well as the berthing maneuver (\Cref{fig:cap_out}).
            The captain, in the first phase, departs from the berthed position with a translational motion.
            In the second phase, the ship is accelerated after taking off to head for the port exit.
            
            These boundary states were set as the checkpoint states because they can be considered as characteristics in the captain maneuvering data.
            They are $\mathcal{C}_{ \mathrm{in}, 2 }$ for berthing, and $\mathcal{C}_{ \mathrm{out}, 1 }$ for unberthing, respectively.
            In the captain maneuvering data, these two checkpoint states are chosen at $t = 572 ~ \mathrm{[s]}$ and $t = 160 ~ \mathrm{[s]}$ in berthing and unberthing time series, respectively.
            However, the motion of the captain berthing trajectory data was almost completely stopped at this time.
            Therefore, at $\mathcal{C}_{ \mathrm{in}, 2 }$, speed $U_{\mathrm{CP, in, 2}}$ and yaw angular velocity $r_{\mathrm{CP, in, 2}}$ were set to be zero.
            For the same reason, at $\mathcal{C}_{ \mathrm{out}, 1 }$, yaw angular velocity $r_{\mathrm{CP, out, 1}}$ was set to be zero.
            
            Other checkpoint states were also set referring to the captain maneuvering data.
            They are the states when the ship cruises in the center area of the port, both for berthing/unberthing.
            These behavioral characteristics also symbolize the maneuver of the captain.
            In the captain maneuvering data, these are states at $t = 220 ~ \mathrm{[s]}$ and $t = 550 ~ \mathrm{[s]}$ in berthing and unberthing time series, respectively.
            
            \Add{
            Our numerical experiments with two checkpoint conditions were sufficient for a discussion of the proposed method's feasibility.
            The second checkpoint was not chosen arbitrarily but rather based on the captain maneuvering data.
            The authors ran a numerical experiment with these settings of checkpoint conditions for one berthing and one unberthing case.
            The outcome was compared to the captain maneuvering data in both cases.
            As a result, numerical experiments can be an important source of debate regarding the feasibility of the proposed algorithm.
            Because an autonomous checkpoint generation algorithm has not yet been developed, checkpoints must be set manually.
            Our research group is currently researching this topic, and we are analyzing a large number of captain maneuvers to understand the characteristics of captain maneuvers and the way human operators think.
            The authors of the current study, on the other hand, did not place a premium on the nature of the checkpoints.
            Rather, the checkpoints were determined using the captain meneuvering data.
            Then, the authors checked whether the proposed algorithm could produce trajectories with the same properties as the captain maneuvering data with these checkpoint conditions.
            }
            
            The checkpoint states and their tolerance values utilized herein are summarized as follows.
            In both the berthing/unberthing trajectory planning cases, two checkpoint conditions were employed.
            \Cref{fig:cap_in,fig:cap_out} show the colored checkpoint states.
            \Cref{tab:CP_setting_in,tab:CP_setting_out} show the checkpoint states and their tolerance values.
            \begin{table}[h]
                \centering
                \caption{Calculation condition on checkpoints $\mathcal{C}_{ \mathrm{in}, 1 }$ and $\mathcal{C}_{ \mathrm{in}, 2 }$ for berthing maneuver optimization.}
                \begin{tabular}{|c||c|c|}
                    \hline
                    Item & Value at $\mathcal{C}_{ \mathrm{in}, 1 }$ & Value at $\mathcal{C}_{ \mathrm{in}, 2 }$  \\
                    \hline
                    \hline
                    $ ( x_{\mathrm{CP}}, y_{\mathrm{CP}} ) $ [m] & $ ( 618, 243 ) $ & $ ( 227, 312 ) $  \\
                    \hline
                    $ d_{\mathrm{CP}, \mathrm{tol}} $ [m] & $ 0.5L_{\mathrm{pp}} $ & $ 0.5L_{\mathrm{pp}} $  \\
                    \hline
                    $ \psi_{\mathrm{CP}} $ [rad] & $ 3.260 $ & $1.155$  \\
                    \hline
                    $ \psi_{\mathrm{CP, tol}} $ [rad] & $ \pi / 180 $ & $ \pi / 180 $  \\
                    \hline
                    $ U_{\mathrm{CP}} $ [m/s] & $3.36$ & $ 0.0 $  \\
                    \hline
                    $ U_{\mathrm{CP, tol}} $ [knots] & $ 0.5 $ & $ 0.5 $  \\
                    \hline
                    $ r_{\mathrm{CP}} $ [rad/s] & $ 7.51 \times 10 ^ {-3} $ & $ 0.0 $  \\
                    \hline
                    $ r_{\mathrm{CP, tol}} $ [rad/s] & $ 9.28 \times 10 ^ {-3} $ & $ 9.28 \times 10 ^ {-3} $  \\
                    \hline
                \end{tabular}
                \label{tab:CP_setting_in}
            \end{table}
            \begin{table}[h]
                \centering
                \caption{Calculation condition on checkpoints $\mathcal{C}_{ \mathrm{out}, 1 }$ and $\mathcal{C}_{ \mathrm{out}, 2 }$ for unberthing maneuver optimization.}
                \begin{tabular}{|c||c|c|}
                    \hline
                    Item & Value at $\mathcal{C}_{ \mathrm{out}, 1 }$ & Value at $\mathcal{C}_{ \mathrm{out}, 2 }$  \\
                    \hline
                    \hline
                    $ ( x_{\mathrm{CP}}, y_{\mathrm{CP}} ) $ [m] & $ ( 5, -69 ) $ & $ ( 668, 179 ) $  \\
                    \hline
                    $ d_{\mathrm{CP}, \mathrm{tol}} $ [m] & $ 0.5L_{\mathrm{pp}} $ & $ 0.5L_{\mathrm{pp}} $  \\
                    \hline
                    $ \psi_{\mathrm{CP}} $ [rad] & $ 1.064 $ & $ 0.830 $  \\
                    \hline
                    $ \psi_{\mathrm{CP, tol}} $ [rad] & $ \pi / 180 $ & $ \pi / 180 $  \\
                    \hline
                    $ U_{\mathrm{CP}} $ [m/s] & $ 0.41 $ & $ 2.74 $  \\
                    \hline
                    $ U_{\mathrm{CP, tol}} $ [knots] & $ 0.5 $ & $ 0.5 $  \\
                    \hline
                    $ r_{\mathrm{CP}} $ [rad/s] & $ 0.0 $ & $ 7.90 \times 10 ^ {-3} $  \\
                    \hline
                    $ r_{\mathrm{CP, tol}} $ [rad/s] & $ 9.28 \times 10 ^ {-3} $ & $ 9.28 \times 10 ^ {-3} $  \\
                    \hline
                \end{tabular}
                \label{tab:CP_setting_out}
            \end{table}
            In \Cref{tab:CP_setting_in,tab:CP_setting_out}, $ d_{\mathrm{CP}, \mathrm{tol}} $ is $ 1 ~ \mathrm{[deg]} $ and $ r_{\mathrm{CP, tol}} $ is the yaw angular velocity as the head and the stern move at $0.5 ~ \mathrm{[m/s]}$.
            Also, note that the units are different in the rows of $ U_{\mathrm{CP}} $ and $ U_{\mathrm{CP, tol}} $.
            
        \subsubsection{Other settings}
        
            \label{sec:calc_cond_other}
            
            Other necessary calculation settings are described in \Cref{sec:calc_cond_other}.
            These settings were determined following the method proposed by \citet{Miyauchi2022}.
            
            The settings for hyperparameters in the objective function of \Cref{eq:J} are shown in \Cref{tab:calc_cond_terminal}.
            \begin{table*}[t]
                \centering
                \caption{Hyperparameters in objective function}
                \begin{tabular}{|c|c|c|}
                    \hline
                    Item & \multicolumn{2}{c|}{Value}  \\
                    \hline
                    \hline
                    & Berthing & Unberthing  \\
                    \hline
                    $\bm{w}_{\mathrm{dim}}$
                    &
                    \multicolumn{2}{c|}{
                        $
                            (
                                1 / w_{L}^{2},
                                1 / w_{U}^{2},
                                1 / w_{L}^{2},
                                1 / w_{U}^{2},
                                \pi^{2},
                                w_{L}^{2} / w_{U}^{2}
                            )
                        $
                    }  \\
                    \hline
                    $\bm{x}_{\mathrm{tol}}$
                    &
                    \begin{tabular}{c}
                        $
                            (
                                1.0 ~ \mathrm{[m]},
                                0.1 ~ \mathrm{[m/s]},
                                1.0 ~ \mathrm{[m]},
                        $  \\
                        $
                                0.1 ~ \mathrm{[m/s]},
                                \Add{\pi / 360 ~ \mathrm{[rad]}},
                                0.1 \times 2 / L_{\mathrm{pp}} ~ \mathrm{[rad/s]}
                            )
                        $
                    \end{tabular}
                    &
                    \begin{tabular}{c}
                        $
                            (
                                1.0 ~ \mathrm{[m]},
                                0.1 ~ \mathrm{[m/s]},
                                1.0 ~ \mathrm{[m]},
                        $  \\
                        $
                                0.1 ~ \mathrm{[m/s]},
                                \pi / 180 ~ \mathrm{[rad]},
                                0.1 \times 2 / L_{\mathrm{pp}} ~ \mathrm{[rad/s]}
                            )
                        $
                    \end{tabular}  \\
                    \hline
                    $w_{\mathrm{C}}$ & \multicolumn{2}{c|}{$10 ^ {6}$}  \\
                    \hline
                    $w_{\mathrm{pen}}$ & \multicolumn{2}{c|}{$10 ^ {4}$}  \\
                    \hline
                \end{tabular}
                \label{tab:calc_cond_terminal}
            \end{table*}

            In \Cref{tab:calc_cond_terminal}, $w_{L} = 0.1 L_{\mathrm{pp}}$ and $ w_{U} = 2.57 $ (the speed at the beginning of berthing maneuver in the captain maneuvering data) for berthing trajectory planning, and $ w_{U} = 2.42 $ (the speed at the completion of unberthing maneuver in the captain maneuvering data) for unberthing trajectory planning were set, respectively.
            
            Using these hyperparameters, the risk of collisions with obstacles, extremely close approaches to obstacles, and break of terminal condition and checkpoint conditions are measured and penalized.
            As described in \Cref{sec:J_before_ship_domain}, the solution without collisions of the ship domain and the area of the surroundings gives $C  = 0$ in the objective function.
            If the solution also satisfies the terminal condition, the second term in \Cref{eq:J_before} becomes $\sum_{i = 1}^{6} w_{\mathrm{dim}, i} x_{\mathrm{tol}, i}^{2}$.
            Of course, with the use of the hyperparameters in \Cref{tab:calc_cond_terminal}, it becomes:
            \begin{equation}
                \sum_{i = 1}^{6} w_{\mathrm{dim}, i} x_{\mathrm{tol}, i}^{2} \simeq 1.08 \times 10 ^ {-2}
                \enspace .
            \end{equation}
            Furthermore, from \Cref{alg:P_CP_pos,alg:P_CP_psi,alg:P_CP_U,alg:P_CP_r}, the penalty value for each mode of the checkpoint condition is within $1$ if each inequality condition of that mode is satisfied.
            For example, if two checkpoint conditions are set, $P_{\mathrm{CP}}$ will contain eight terms, and we know that $P_{\mathrm{CP}} \leq 8$ if all checkpoint conditions are satisfied.
            Therefore, if the optimization results with the appropriate value of objective function are obtained, it can be judged that the solution 
            (a) satisfies all checkpoint conditions,
            (b) satisfies the terminal condition,
            (c) and realizes the trajectory without collision and interference of the ship domain.
            For example, for results which is about $t_{\mathrm{f}} \simeq 1.0 \times 10 ^ {3} ~ \mathrm{[s]}$ long, such as captain maneuvering data, appropriate value is about $J \simeq t_{\mathrm{f}} \times 10 ^ {-2}$.
            In addition, the obtained solutions that give larger values, such as greater degrees of order, can be considered to be affected by $w_{\mathrm{C}}$ or $w_{\mathrm{pen}}$ or $w_{\mathrm{CP, pen}}$.
            In other words, such solutions contain the intrusion of the ship domain or do not satisfy the terminal conditions and checkpoint conditions.
            
            For both berthing and unberthing trajectory plannings, the initial and target terminal states were the starting and ending states in the captain maneuvering data, respectively.
            However, the velocities and yaw angular velocities of the terminal condition were set to be zero for the berthing trajectory planning.
            Similarly, the velocities and yaw angular velocities for the initial condition were set to be zero for the unberthing trajectory planning.
            The wind condition was set by considering the representative value shown in \Cref{tab:cap_data} as the steady wind.
            
            The search ranges of the terminal time and instantaneous control inputs are summarized in \Cref{tab:exploring_range}.
            \begin{table}[tb]
                \centering
                \caption{Exploring range of parameters to optimize.}
                \label{tab:exploring_range}
                \begin{tabular}{|c|c|}
                    \hline
                    Variable & Exploring limit  \\
                    \hline
                    \hline
                    $t_{\mathrm{f}} ~ \mathrm{[s]}$ (for berthing) & $[ 900, 1800 ]$  \\
                    \hline
                    $t_{\mathrm{f}} ~ \mathrm{[s]}$ (for unberthing) & $[ 200, 1080 ]$  \\
                    \hline
                    Rudder angle [deg] & $[-35, 35]$  \\
                    \hline
                    \begin{tabular}{c}
                        Propeller revolution [rps]  \\
                        (for berthing)
                    \end{tabular}
                    &
                    $[-1.0, 1.0]$  \\
                    \hline
                    \begin{tabular}{c}
                        Propeller revolution [rps]  \\
                        (for unberthing)
                    \end{tabular}
                    &
                    $[0.0, 1.0]$  \\
                    \hline
                    Bow thruster revolution [rps] & $[-5.18, 5.18]$  \\
                    \hline
                    Stern thruster revolution [rps] & $[-6.22, 6.22]$  \\
                    \hline
                \end{tabular}
            \end{table}
            The terminal time was determined by referring to the maneuvering time in the captain maneuvering data.
            The search range of the instantaneous control inputs was defined from the mechanical limits of the subject ship.
            
            The hyperparameters concerning the shape of the elliptical ship domain are shown in \Cref{tab:ship_domain_parameter}.
            \begin{table}[h]
                \centering
                \caption{Hyperparameters which determine elliptical ship domain in proposed algorithm.}
                \begin{tabular}{|c|c|}
                    \hline
                    Hyperparameter & Value  \\
                    \hline
                    \hline
                    $U_{\mathrm{max}}$ & $6.0 ~ \mathrm{[knots]}$  \\
                    \hline
                    $U_{\mathrm{min}}$ & $2.0 ~ \mathrm{[knots]}$  \\
                    \hline
                    $L_{\mathrm{longi, max, long}}$ & $0.85 L_{\mathrm{pp}}$  \\
                    \hline
                    $L_{\mathrm{longi, max, short}}$ & $0.5 L_{\mathrm{pp}}$  \\
                    \hline
                    $L_{\mathrm{longi, min}}$ & $0.5 L_{\mathrm{pp}}$  \\
                    \hline
                    $L_{\mathrm{lat, max}}$ & $0.39 L_{\mathrm{pp}}$  \\
                    \hline
                    $L_{\mathrm{lat, min}}$ & $0.18 L_{\mathrm{pp}}$  \\
                    \hline
                \end{tabular}
                \label{tab:ship_domain_parameter}
            \end{table}
            Also, an overview of the elliptical ship domain is shown in \Cref{fig:tear_drop_domain}.
            The hyperparameters defining the elliptical ship domain are determined based on the method proposed by \citet{Miyauchi2022}, as described in \Cref{sec:J_before_ship_domain}.
            These hyperparameters were set considering the minimum passage width $W \simeq L_{\mathrm{pp}}$ in the target port.
            However, following this method, the maximum margin length for the shorter longitudinal direction $L_{\mathrm{longi, max, short}}$ became smaller than the minimum margin length for this direction $L_{\mathrm{longi, min}}$.
            Therefore, the margin length for the shorter longitudinal direction was set to be constant at $0.5 L_{\mathrm{pp}}$, which is the minimum value.
            The shape of the rectangular ship domain adopted near the target berthed position and the switching between the two regions are described in \Cref{sec:region_switch}.

    \subsection{Results}\label{sec:results}
    
        First, the values of objective function $J$ for the variables $\bm{X}$ optimized by the berthing/unberthing trajectory planning are shown in \Cref{tab:eval_values}.
        \Add{Both calculations were completed within a few tens of hours.}
        \begin{table}[]
            \centering
            \caption{Values of objective function $J$ calculated on optimized variables $\bm{X}$ for berthing and unberthing.}
            \begin{tabular}{|c||c|c|}
                \hline
                Trajectory type & Berthing & Unberthing  \\
                \hline
                \hline
                $J$ (\Cref{eq:J}) & \Add{$8.78$} & $6.47$  \\
                \hline
            \end{tabular}
            \label{tab:eval_values}
        \end{table}
        Considering the arguments mentioned in \Cref{sec:calc_cond_other}, we can confirm that the values shown in \Cref{tab:eval_values} indicate that the optimized trajectories successfully satisfy all the terminal and checkpoint conditions without collisions with surrounding obstacles.
        
\begin{figure*}
            \centering
            \includegraphics[width=1.0\hsize]{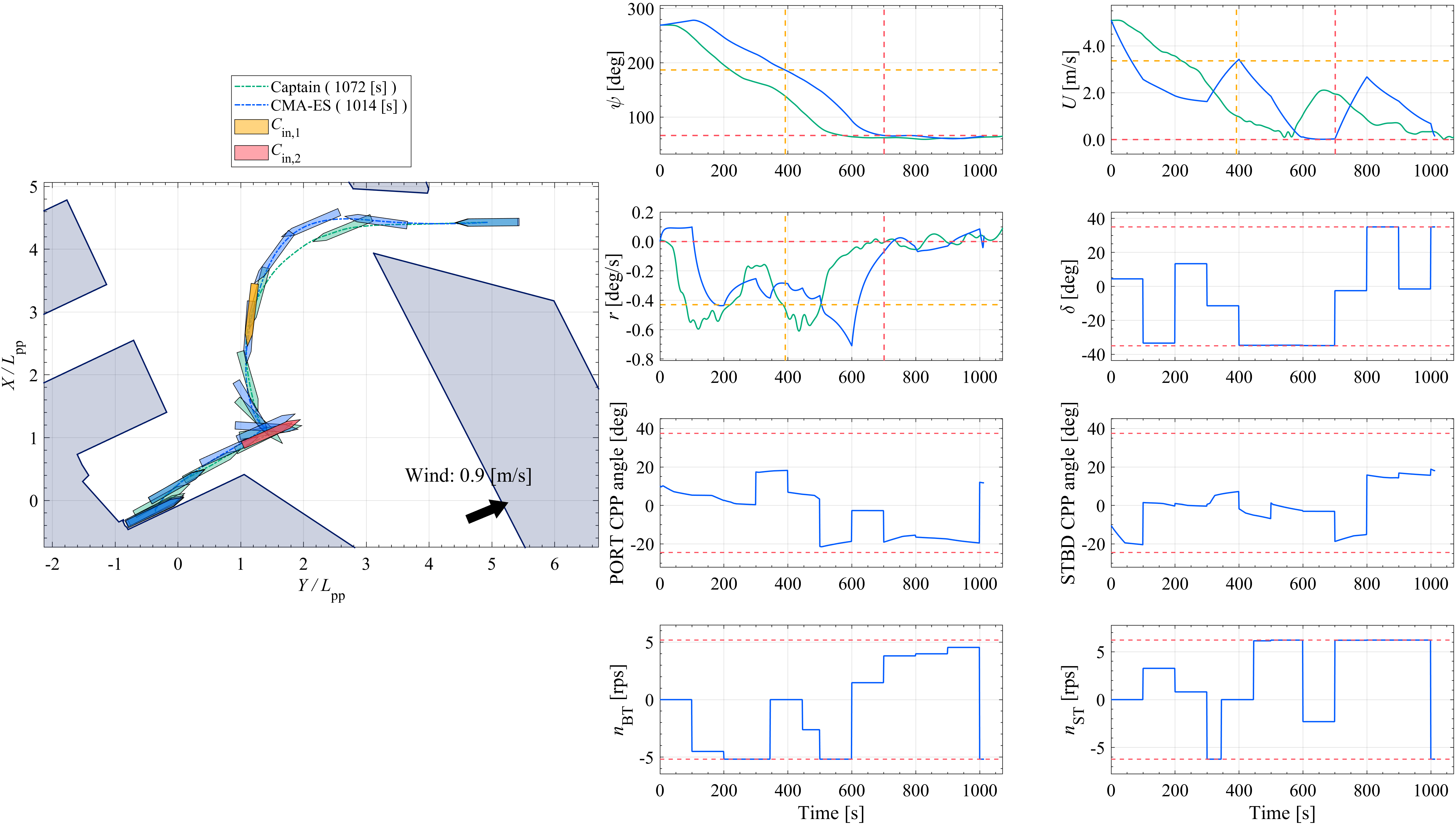}
            \caption{Result for berthing trajectory. Blue-colored data is optimized by the proposed algorithm with checkpoints $\mathcal{C}_{ \mathrm{in}, 1 }$ and $\mathcal{C}_{ \mathrm{in}, 2 }$. Green-colored data presents captain maneuvering data.}
            \label{fig:compari_in}
        \end{figure*}

\begin{figure*}
            \centering
            \includegraphics[width=1.0\hsize]{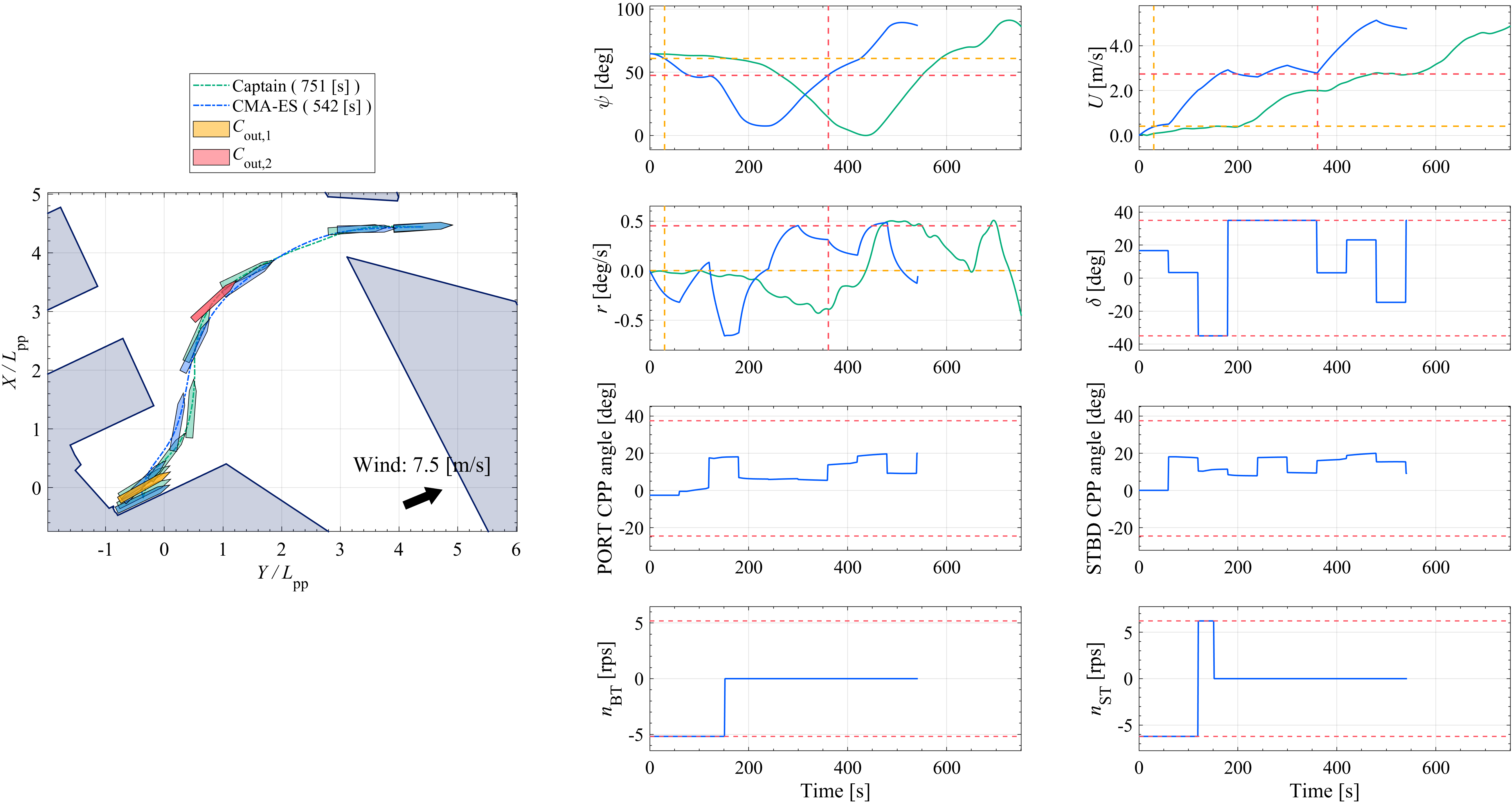}
            \caption{Result for unberthing trajectory. Blue-colored data is optimized by the proposed algorithm with checkpoints $\mathcal{C}_{ \mathrm{out}, 1 }$ and $\mathcal{C}_{ \mathrm{out}, 2 }$. Green-colored data presents captain maneuvering data.}
            \label{fig:compari_out}
        \end{figure*}

        The time series of control inputs and states optimized by the proposed algorithm are shown in \Cref{fig:compari_in,fig:compari_out} for the berthing and unberthing trajectories.
        In \Cref{fig:compari_in,fig:compari_out}, the captain maneuvering data is also shown for comparison.
        \Cref{fig:compari_in,fig:compari_out} show the entire trajectory on the left side and the time series of the state and control inputs on the other graphs.
        The data shown in blue and green are the optimized results and the captain maneuvering data, respectively.
        The trajectory diagrams show the checkpoint states $(x_{\mathrm{CP}}, y_{\mathrm{CP}})$ and $\psi_{\mathrm{CP}}$ in different colored ship figures.
        The time series of $\psi$, $U$, and $r$ also show information about the checkpoint conditions with dotted lines.
        The horizontal line represents the checkpoint state and the vertical line represents the time when $P_{\mathrm{CP}, i}$ defined by \Cref{eq:P_CP_i} is minimized.
        Concerning the berthing trajectory, the times of minimum penalties for $\mathcal{C}_{ \mathrm{in}, 1 }$ and $\mathcal{C}_{ \mathrm{in}, 2 }$ were $t = \Add{392} ~ \mathrm{[s]}$ and $t = \Add{701} ~ \mathrm{[s]}$, respectively.
        Concerning the unberthing trajectory, the times of minimum penalties for $\mathcal{C}_{ \mathrm{out}, 1 }$ and $\mathcal{C}_{ \mathrm{out}, 2 }$ were $t = 30 ~ \mathrm{[s]}$ and $t = 361 ~ \mathrm{[s]}$, respectively.
        Since the captain maneuvering data precludes the time series of control inputs, the graphs of the control inputs show only the optimization results.
        As stated previously, the CPP pitch angles are converted from the FPPs revolution number of the optimization results.
        
        \Cref{fig:compari_in,fig:compari_out} also show that, in the time series of state variables, they take close values to the given checkpoint states simultaneously at a certain time.
        In fact, the acquired trajectories were found to satisfy all checkpoint conditions.
        \Cref{tab:actual_diff} shows the satisfaction of the checkpoint conditions in the results.
        \begin{table*}[]
            \centering
            \caption{Deviations between actual states in results and designated checkpoint states and tolerance values for every four modes of the checkpoint conditions.}
            \begin{tabular}{|c||c|c|c|c|c|}
                \hline
                \multirow{3}{*}{} & \multicolumn{4}{c|}{Deviation} & \multirow{3}{*}{Tolerance}  \\
                \cline{2-5}
                & \multicolumn{2}{c|}{Berthing} & \multicolumn{2}{c|}{Unberthing} &  \\
                \cline{2-5}
                &
                \begin{tabular}{c}
                    $t = 403 ~ \mathrm{[s]}$  \\
                    (for $\mathcal{C}_{ \mathrm{in}, 1 }$)
                \end{tabular}
                &
                \begin{tabular}{c}
                    $t = 701 ~ \mathrm{[s]}$  \\
                    (for $\mathcal{C}_{ \mathrm{in}, 2 }$)
                \end{tabular}
                &
                \begin{tabular}{c}
                    $t = 30 ~ \mathrm{[s]}$  \\
                    (for $\mathcal{C}_{ \mathrm{out}, 1 }$)
                \end{tabular}
                &
                \begin{tabular}{c}
                    $t = 361 ~ \mathrm{[s]}$  \\
                    (for $\mathcal{C}_{ \mathrm{out}, 2 }$)
                \end{tabular}
                &  \\
                \hline
                \hline
                Position $\mathrm{[m]}$
                    & $\Add{0.284} \times 10 ^ {2}$
                    & $0.224 \times 10 ^ {2}$
                    & $\Add{0.258} \times 10 ^ {2}$
                    & $0.128 \times 10 ^ {2}$
                    & $1.047 \times 10 ^ {2}$  \\
                \hline
                $\psi ~ \mathrm{[rad]}$
                    & $\Add{0.000} \times 10 ^ {-2}$
                    & $0.029 \times 10 ^ {-2}$
                    & $\Add{0.008} \times 10 ^ {-2}$
                    & $0.007 \times 10 ^ {-2}$
                    & $1.745 \times 10 ^ {-2}$  \\
                \hline
                $U ~ \mathrm{[m/s]}$
                    & $\Add{0.484} \times 10 ^ {-1}$
                    & $0.394 \times 10 ^ {-1}$
                    & $\Add{0.285} \times 10 ^ {-1}$
                    & $0.026 \times 10 ^ {-1}$
                    & $2.572 \times 10 ^ {-1}$  \\
                \hline
                $r ~ \mathrm{[rad/s]}$
                    & $\Add{2.546} \times 10 ^ {-3}$
                    & $2.434 \times 10 ^ {-3}$
                    & $\Add{1.225} \times 10 ^ {-3}$
                    & $0.805 \times 10 ^ {-3}$
                    & $9.287 \times 10 ^ {-3}$  \\
                \hline
            \end{tabular}
            \label{tab:actual_diff}
        \end{table*}
        From \Cref{tab:actual_diff}, it is obvious that each of the deviations between the states in the results and the checkpoint states is less than the respective tolerance.
        
        \Cref{fig:compari_in,fig:compari_out} show that, notably, the ship's paths in berthing/unberthing trajectories in the results almost overlap that in the captain maneuvering data.
        The following characteristics can be identified for the two paths from \Cref{fig:compari_in}.
        \Add{
        After passing through the port entrance, the path of the captain maneuvering data first makes a left turn and approaches the checkpoint $\mathcal{C}_{ \mathrm{in}, 1 }$.
        The optimized path turns slightly right at the port entrance, then turns left to approach $\mathcal{C}_{ \mathrm{in}, 1 }$, similarly to the captain's path.
        }
        Then, after passing $\mathcal{C}_{ \mathrm{in}, 1 }$, they turn left and approach $\mathcal{C}_{ \mathrm{in}, 2 }$ on a course orthogonal to the checkpoint heading at $\mathcal{C}_{ \mathrm{in}, 2 }$.
        After stopping at the checkpoint $\mathcal{C}_{ \mathrm{in}, 2 }$ ($U_{\mathrm{CP}} = 0 ~ \mathrm{[m/s]}$), the ship moves back to the target berthed position in a translational motion.
        \Cref{fig:compari_out} also shows the following features in unberthing trajectories.
        Both trajectories move in a parallel motion to the north after the start of the maneuver, and then leave the berth on the starboard side.
        Afterward, the two paths turn right, going to the north, and move on toward the port exit.
        For the first half of the unberthing trajectories, the path in the result was more linear than that of the captain maneuvering data, although it satisfied the checkpoint condition.
        However, an overall similar behavior match between the planned trajectory and captain maneuvering data was not observed in the time series of ship speed $U$ and yaw angular velocity $r$.
        
\begin{figure*}
            \centering
            \includegraphics[width=1.0\hsize]{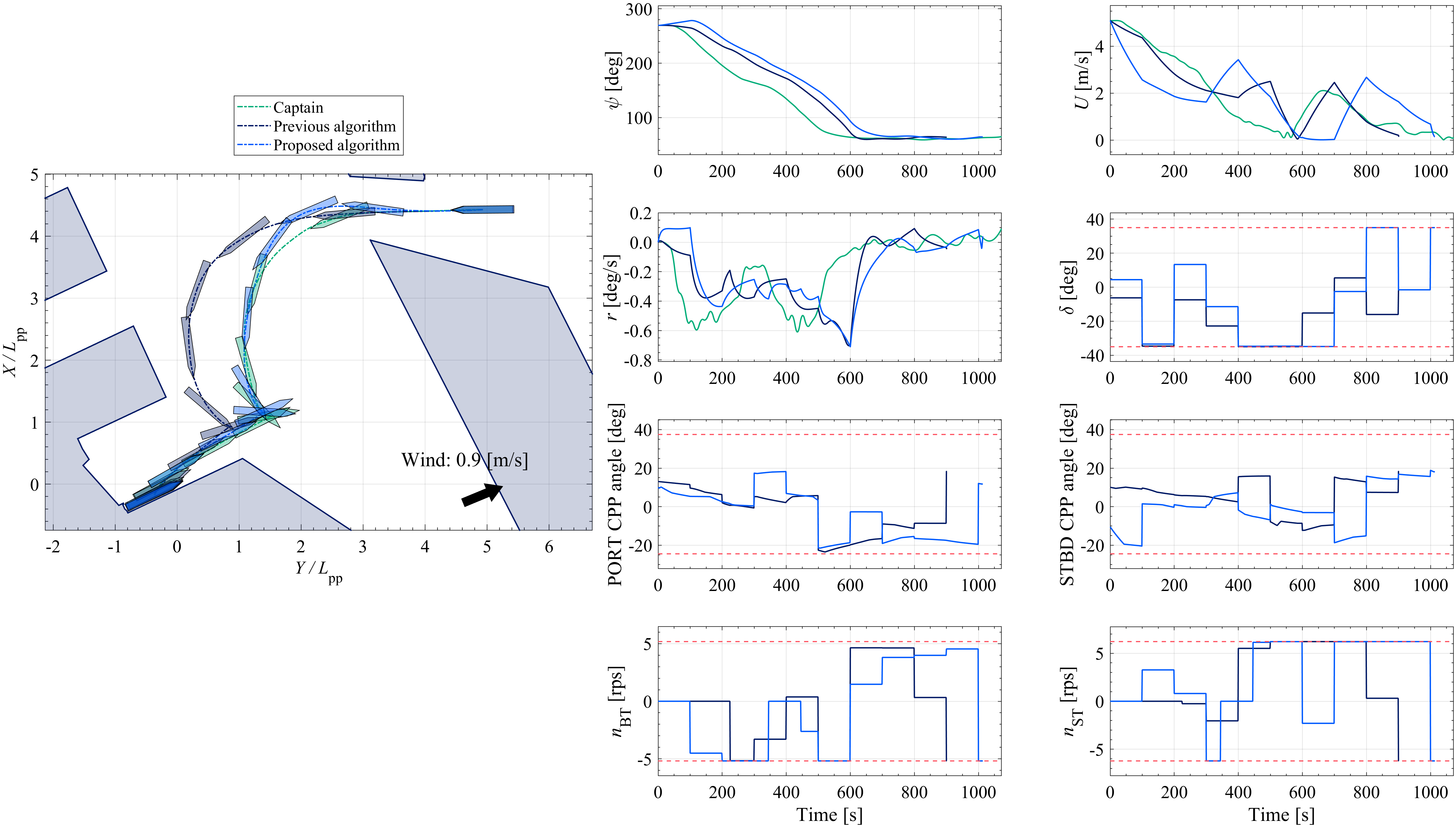}
            \caption{Comparison between three trajectories. Blue-colored data is optimized with the proposed algorithm with checkpoints $\mathcal{C}_{ \mathrm{in}, 1 }$ and $\mathcal{C}_{ \mathrm{in}, 2 }$. Navy-colored data is optimized with the previous algorithm reported by \citet{Miyauchi2022} which does not consider the checkpoint condition. Green-colored data is the captain maneuvering data.}
            \label{fig:compari_in_prev}
        \end{figure*}

\begin{figure*}
            \centering
            \includegraphics[width=1.0\hsize]{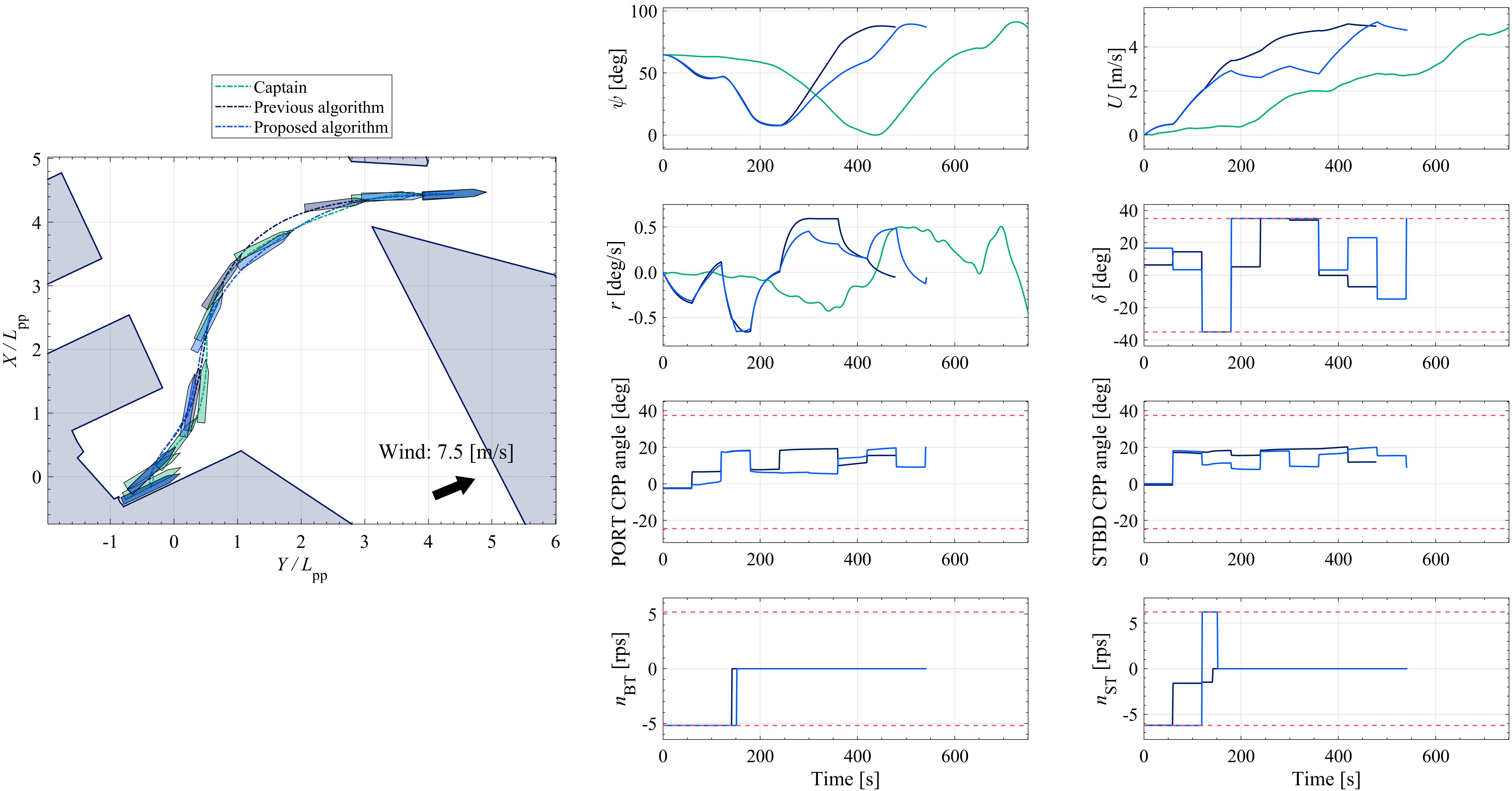}
            \caption{Comparison between three trajectories. Blue-colored data is optimized with the proposed algorithm with checkpoints $\mathcal{C}_{ \mathrm{out}, 1 }$ and $\mathcal{C}_{ \mathrm{out}, 2 }$. Navy-colored data is optimized with the previous algorithm reported by \cite{Miyauchi2022} which does not consider the checkpoint condition. Green-colored data is the captain maneuvering data.}
            \label{fig:compari_out_prev}
        \end{figure*}

        The results obtained herein were also compared with that from the previous algorithm.
        To do so, we performed calculations using the previous method \citep{Miyauchi2022} that does not consider the checkpoint conditions under the same calculation conditions. The comparisons are shown in \Cref{fig:compari_in_prev,fig:compari_out_prev}.

        In \Cref{fig:compari_in_prev,fig:compari_out_prev}, the data shown in blue are the results optimized by the proposed algorithm.
        The navy-colored data is the optimized results under the same conditions using the previous method \citep{Miyauchi2022}.
        The green-colored data is the captain maneuvering data.
        From \Cref{fig:compari_in_prev,fig:compari_out_prev}, it can be confirmed that the ship's paths generated by the proposed algorithm resemble the captain maneuvering data for both berthing and unberthing much.
        \Cref{fig:compari_in_prev} indicates that the blue and green-colored paths almost overlap, as described in the previous paragraph.
        In contrast, the navy-colored path moves westward from the other two paths after the control start and takes a course closer to the obstacle.
        Then, the navy-colored path adjusts its heading and shifts to the backward motion before approaching the target berthed position by translational motion.
        However, the position of its large turn is also westward.
        In \Cref{fig:compari_out_prev}, showing the unberthing trajectory, the difference between the algorithms did not appear significant.
        The three paths proceed in the same manner until around the northern checkpoint $\mathcal{C}_{ \mathrm{out}, 2 }$ (\Cref{fig:compari_out}).
        However, the paths after passing the position of $\mathcal{C}_{ \mathrm{out}, 2 }$ differed slightly among the three paths.
        Then, the navy-colored path extends west slightly to the north while navigating to the port exit.
        However, in the result from the proposed algorithm, the path goes straight to the port exit similar to the captain maneuvering data.
        Consequently, the two paths, the result of the new method and the captain maneuvering data, almost overlap in the region $1 \leq Y_{0} / L_{\mathrm{pp}} \leq 5$.
        Thus, concerning the unberthing maneuver, the three trajectories are generally similar, but the proposed method outputs a trajectory with a slightly closer ship path to the actual trajectory by the captain.

\section{Discussion}

    In \Cref{sec:num_calc}, the comparison between the proposed and previous algorithms results showed that the former resembles the captain maneuvering data much.
    The algorithm proposed herein does not impose strong constraints on the state, for example, constraints over the entire time domain.
    Therefore, the time series of states such as $U$ and $r$ in the generated trajectories will not necessarily have the same overall behavior as the captain maneuvering data.
    However, note that even though only a few state constraints were imposed, the optimized ship paths generally overlapped with those of the captain maneuvering data, \Add{indicating that the proposed algorithm is feasible.}
    
    The results illustrated that our new method completely takes over the advantages of the previous method.
    We formulated a trajectory planning algorithm that satisfies the state inequality condition by adding a penalty term to the objective function of the existing method.
    In the proposed method, the optimization landscape of the problem could be more complex than those in previous studies due to the inclusion of newly added terms.
    Nevertheless, numerical experiments showed that the proposed algorithm successfully generated solutions that satisfy (a) all checkpoint conditions, (b) noninterference between the ship domain and the structure domains, (c) and the tolerated terminal condition.
    The results show that the formulated new method does not undermine the advantages of the previous report.
    
    It is thought that trajectory planning by the proposed algorithm can be used not only as a recommended trajectory, but also as a reference trajectory for autonomous ship berthing/unberthing control.
    Considering the commercial scene of autonomous ship operation, there exist stakeholders, such as shipowners and shipping companies.
    Here, they prefer traditional human-like operations because their financial activities are cost-intensive and are accustomed to control characteristics of human operators.
    In this situation, it would be reassuring to them if they could explicitly direct the ship's motion corresponding to be customary maneuver in the autonomous operation of ships.
    Therefore, our newly proposed algorithm is applicable to generating reference trajectories for ship autonomy.
    
    There are several limitations of the proposed algorithm.
    As pointed out in the previous report, trajectory planning calculations using CMA-ES take several hours to days.
    Thus, for the ship operator to use the plan as a reference, it must be prepared considering the calculation time.
    However, since CMA-ES computation is suitable for parallelized calculation, the CMA-ES computation time can be shortened by introducing parallel computation.
    Another limitation is that the trajectory plan generated by the proposed algorithm depends on the considered checkpoint conditions.
    This study did not address the best practice of setting the checkpoint conditions to generate a recommended trajectory.
    However, a captain with some experience can set the checkpoint conditions based on knowledge and experience of ship maneuvers.
    At this stage, if checkpoint conditions are challenging to set, the captain may establish the checkpoint conditions by seeking direction from a skilled senior captain or pilot in advance.
    The method of autonomously setting the checkpoint conditions is a potential future work.

\section{Conclusion}
    
    A gentle state constraint named ``checkpoint condition'' was newly introduced in the ship berthing/unberthing trajectory planning problem.
    The checkpoint condition is novel in that it can impose mild inequality constraints not only on position, but also on heading, ship speed, and yaw angular velocity during ship motion.
    Furthermore, we proposed an algorithm to generate the trajectory that simultaneously satisfies these four conditions: position, heading, speed, and yaw angular velocity, at a certain time.
    In our method, the checkpoint condition is treated in the objective function using the penalty method.
    Using the proposed algorithm, it could be possible to impose a condition, indicating that states are approximately realized at a certain time in the trajectory plans.
    
    The numerical experimental results showed that the ship paths generated by the proposed method are almost identical to the paths of the actual operational data.
    This is notable because the additional condition, compared to the previous work, is just that the state should be close to the indicated value at a certain time.
    We also confirmed that the generated trajectory by the proposed algorithm satisfies all the checkpoint conditions and completely takes over the advantages of previous work.
    
    The proposed algorithm can generate a recommended trajectory because careful maneuver, as a human captain would do, can be considered in our algorithm.
    Generally, the recommended trajectory should have the control characteristics of the human operator.
    The proposed algorithm enables planning a berthing/unberthing trajectory with such requests by solving the optimal control problem with inequality constraints on states.


\printcredits


\section*{Acknowledgment}

    This study was conducted jointly with Mitsubishi Heavy Industries, Ltd. with operational data and the maneuvering ship motion model of the subject ship provided by them.
    This study was also conducted as part of the Nippon Foundation to Support Unmanned Ship Development Projects ``MEGURI2040''.
    Additionally, this study was supported by a Grant-in-Aid for Scientific Research from the Japan Society for Promotion of Science (JSPS KAKENHI Grant \#19K04858 and \#20H02398).

\appendix
\section{\Add{Terminal state for berthing trajectory}}

    \label{sec:appendix}
    \Add{
    The explanation of the terminal state for the berthing trajectory, which is depicted in \Cref{fig:overview_rec_and_tol,fig:exact_rec_and_tol}, and the validation of our setting for berthing trajectory planning are provided here.
    }
    
    \Add{
    The target terminal state for berthing trajectory was set so that the clearance between the starboard side and berth line was $D = 4.0 ~ \mathrm{[m]}$ at the midship,
    and the heading angle is $\psi_{\mathrm{b}} = 25.10 ~ \mathrm{[deg]}$ from the positive side of $Y$ axis.
    The berth line and the positive side of $Y$ axis makes an angle of $\theta_{\mathrm{b}} = 25.37 ~ \mathrm{[deg]}$.
    Herein, the tolerance for $X$ and $Y$ were $x_{\mathrm{tol}, 1} = x_{\mathrm{tol}, 3} = 1.0 ~ \mathrm{[m]}$ as shown in \Cref{tab:calc_cond_terminal}, and the ship length was $L = 222.5 \mathrm{[m]}$.
    Suppose the terminal condition is satisfied, the closest point of the rectangular ship domain at the terminal state appears when:
    \begin{equation}
        \left \{
            \begin{aligned}
                x_{1}(t_{\mathrm{f}}) &= x_{\mathrm{fin}, 1} + x_{\mathrm{tol}, 1}  \\
                x_{3}(t_{\mathrm{f}}) &= x_{\mathrm{fin}, 3} + x_{\mathrm{tol}, 3}  \\
                x_{5}(t_{\mathrm{f}}) &= x_{\mathrm{fin}, 5} + x_{\mathrm{tol}, 5}
                    \quad ( \because \quad \psi_{\mathrm{b}} < \theta_{\mathrm{b}} )
            \end{aligned}
        \right .
        \enspace .
        \tag{A.1}
    \end{equation}
    Let us denoting the deviation of fore starboard side corner of rectangular ship domain from the target terminal state to deviated state due to $(x_{\mathrm{tol}, 1}, x_{\mathrm{tol}, 3})$ and $x_{\mathrm{tol}, 5}$ as $\Delta_{xy}$ and $\Delta_{\psi}$, respectively
    Then, the following inequation must be satisfied to prevent the intrusion of the berth region:
    \begin{equation}
        \label{eq:ineq}
        D \geq l + \Delta_{xy} + \Delta_{\psi}
        \enspace .
        \tag{A.2}
    \end{equation}
    From a geometric relationship, each deviation is formulated as follows:
    \begin{equation}
        \label{eq:Delta_xy}
        \Delta_{xy} = \sqrt{{x_{\mathrm{tol}, 1}} ^ {2} + {x_{\mathrm{tol}, 3}} ^ {2}} \cos \theta_{\mathrm{m}}
        \enspace ,
        \tag{A.3}
    \end{equation}
    \begin{equation}
        \label{eq:Delta_psi}
        \Delta_{\psi} = \left ( \frac{L}{2} + l \right ) \sin (x_{\mathrm{tol}, 5} + \Tilde{\psi})
        \enspace ,
        \tag{A.4}
    \end{equation}
    where $\theta_{\mathrm{m}} = \pi / 4 - \theta_{\mathrm{b}}$ and $\Tilde{\psi} = \theta_{\mathrm{b}} - \psi_{\mathrm{b}}$.
    Resolving \Cref{eq:ineq} with \Cref{eq:Delta_xy,eq:Delta_psi}, we obtain:
    \begin{equation}
        \label{eq:psi_tol_max}
        x_{\mathrm{tol}, 5}
            \leq
            \arcsin{
                \left (
                    \frac{D - l - \Delta_{xy}}{\frac{L}{2} + l}
                \right )
            }
            -
            \Tilde{\psi}
            \simeq
            0.5799 ~ \mathrm{[deg]}
        \enspace .
        \tag{A.5}
    \end{equation}
    \Cref{eq:psi_tol_max} indicates the maximum value of $x_{\mathrm{tol}, 5}$.
    Our setting $x_{\mathrm{tol}, 5} = 0.5 \mathrm{[deg]}$, which shown in \Cref{tab:calc_cond_terminal}, satisfies \Cref{eq:psi_tol_max}.
    }
    
\appendix

\end{document}